\pdfoutput=1

\documentclass[11pt]{article}

\usepackage[final]{acl}

\usepackage{times}
\usepackage{latexsym}

\usepackage[T1]{fontenc}

\usepackage[utf8]{inputenc}

\usepackage{microtype}

\usepackage{inconsolata}

\usepackage{graphicx}

\usepackage{multirow}
\usepackage{times}
\usepackage{soul}
\usepackage{url}
\usepackage[utf8]{inputenc}
\usepackage{caption}
\usepackage{graphicx}
\usepackage{amsmath}
\usepackage{amsthm}
\usepackage{booktabs}
\usepackage{algorithm}
\usepackage{algorithmic}
\usepackage[switch]{lineno}
\usepackage{multicol}
\usepackage{subcaption}
\usepackage{arydshln}
\usepackage[normalem]{ulem}
\usepackage{caption}
\usepackage{colortbl}
\usepackage{tcolorbox}
\usepackage{enumitem}
\usepackage{marvosym}
\newcommand{\nosection}[1]{\vspace{4.5pt}\noindent\textbf{#1}}
\title{Take Off the Training Wheels! \\Progressive In-Context Learning for Effective  Alignment}

\author{
    \begin{tabular}{c}
  Zhenyu Liu$^1$, Dongfang Li$^1$, Xinshuo Hu$^1$,  \\
    Xinping Zhao$^1$, Yibin Chen$^{2}$, \textbf{Baotian Hu$^{1}$\textsuperscript{\Letter}\thanks{\textsuperscript{\Letter}Corresponding author.},} Min Zhang$^1$ \vspace{.5mm} \\
    \end{tabular}
    \\ \vspace{.5mm}
    \begin{tabular}{c}
    $^1$Harbin Institute of Technology (Shenzhen), Shenzhen, China \\
    $^2$Huawei Cloud, Huawei Technologies Ltd. \\
    \texttt{liuzhenyuhit@gmail.com}, \texttt{chenyibin4@huawei.com}, \\
    \texttt{\{lidongfang,hubaotian,zhangmin2021\}@hit.edu.cn} 
    \end{tabular}
     \\ \vspace{.5mm}
}

\makeatletter
\def\thanks#1{\protected@xdef\@thanks{\@thanks
        \protect\footnotetext{#1}}}
\makeatother

\begin{document}
\maketitle
\begin{abstract}

Recent studies have explored the working mechanisms of In-Context Learning (ICL). However, they mainly focus on classification and simple generation tasks, limiting their broader application to more complex generation tasks in practice.
To address this gap, we investigate the impact of demonstrations on token representations within the practical alignment tasks. We find that the transformer embeds the task function learned from demonstrations into the separator token representation, which plays an important role in the generation of prior response tokens. Once the prior response tokens are determined, the demonstrations become redundant.
Motivated by this finding, we propose an efficient Progressive In-Context Alignment (\textsc{\textbf{PICA}}) method consisting of two stages. In the first few-shot stage, the model generates several prior response tokens via standard ICL while concurrently extracting the ICL vector that stores the task function from the separator token representation. In the following zero-shot stage, this ICL vector guides the model to generate responses without further demonstrations.
Extensive experiments demonstrate that our  \textsc{PICA} not only surpasses vanilla ICL but also achieves comparable performance to other alignment tuning methods. The proposed training-free method reduces the time cost (e.g., 5.45×) with improved alignment performance (e.g., 6.57+). 
Consequently, our work highlights the application of ICL for alignment and calls for a deeper understanding of ICL for complex generations.
The code will be available at \url{https://github.com/HITsz-TMG/PICA}.
\end{abstract}

\section{Introduction}


In-Context Learning (ICL) has attracted growing attention alongside the scaling of Large Language Models (LLMs)~\citep{gpt3}. By conditioning on a handful of input-label pairs as examples, LLMs achieve notable improvements and produce impressive few-shot performance across a range of downstream  tasks~\citep{DBLP:journals/tmlr/WeiTBRZBYBZMCHVLDF22}. After that, numerous studies have explored the working mechanism of ICL and propose several effective methods to enhance ICL~\cite{icl_vector,function_vector,DBLP:conf/emnlp/WangLDCZMZS23,state_vector}. 

However, these works mainly focus on classification tasks and simple generation tasks, which limits the exploration of these methods in more complex generation tasks, such as aligning LLMs with human preferences. As a complex and practical task, alignment typically requires training the model, such as Supervise Fine-Tuning (SFT)~\cite{DBLP:conf/nips/ZhouLX0SMMEYYZG23} and Reinforcement Learning from Human Feedback (RLHF)~\cite{DBLP:conf/nips/Ouyang0JAWMZASR22}. A recent work~\cite{unlock_spell} proposed URIAL, a simple method using in-context examples to align several powerful base LLMs and achieves notable instruction-following performance. The success of URIAL demonstrates the feasibility of in-context alignment and encourages us to explore and optimize ICL in the alignment task.

\begin{figure*}[htbp!]
\centering
\begin{subfigure}[b]{0.24\linewidth}
    \includegraphics[width=\linewidth]{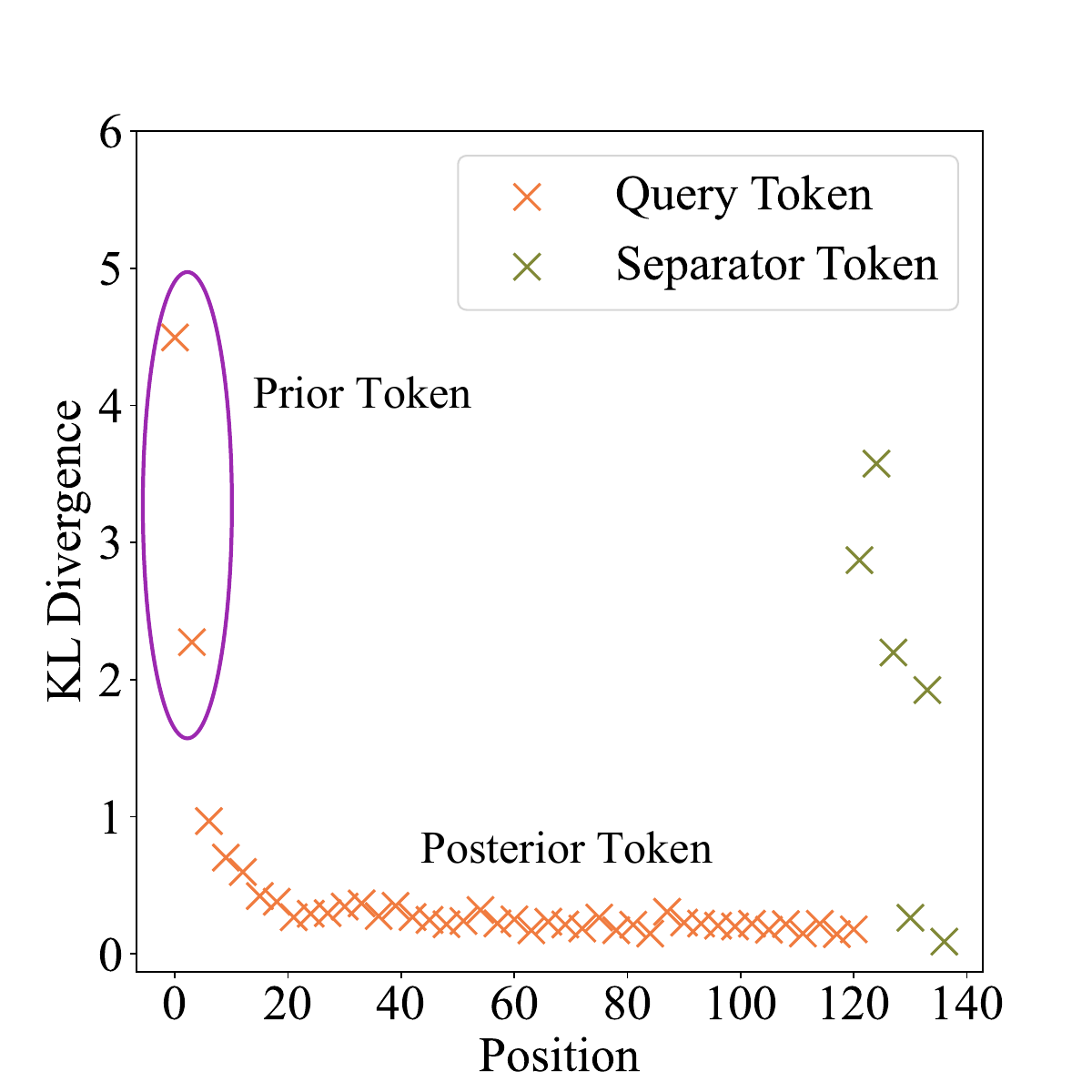}
    \label{subfig:llama_input_exp}
    \caption{Input Experimental Group}
  \end{subfigure}
  \begin{subfigure}[b]{0.24\linewidth}
    \includegraphics[width=\linewidth]{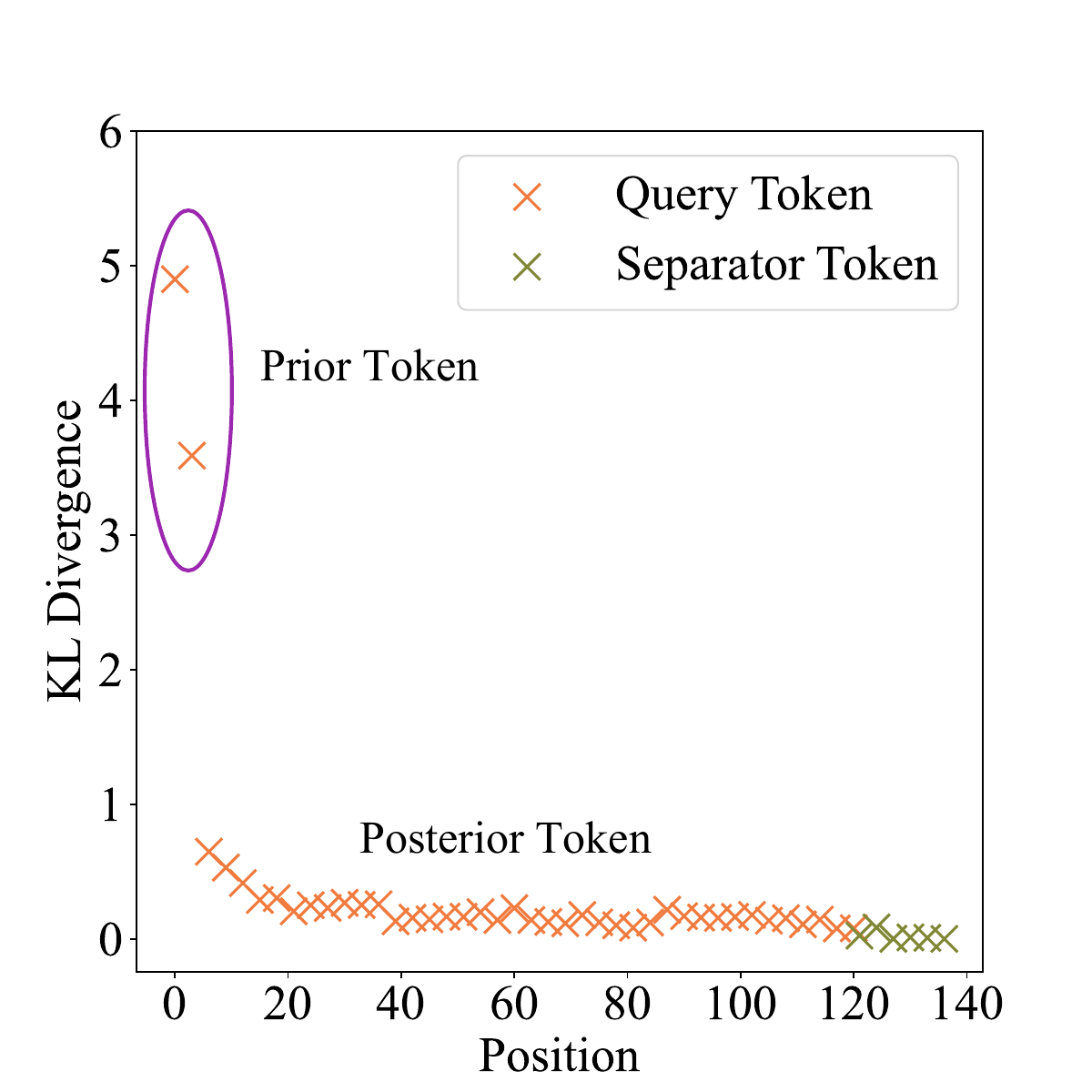}
    \label{subfig:llama_input_con}
    \caption{Input Control Group}
  \end{subfigure}
  \begin{subfigure}[b]{0.24\linewidth}
    \includegraphics[width=\linewidth]{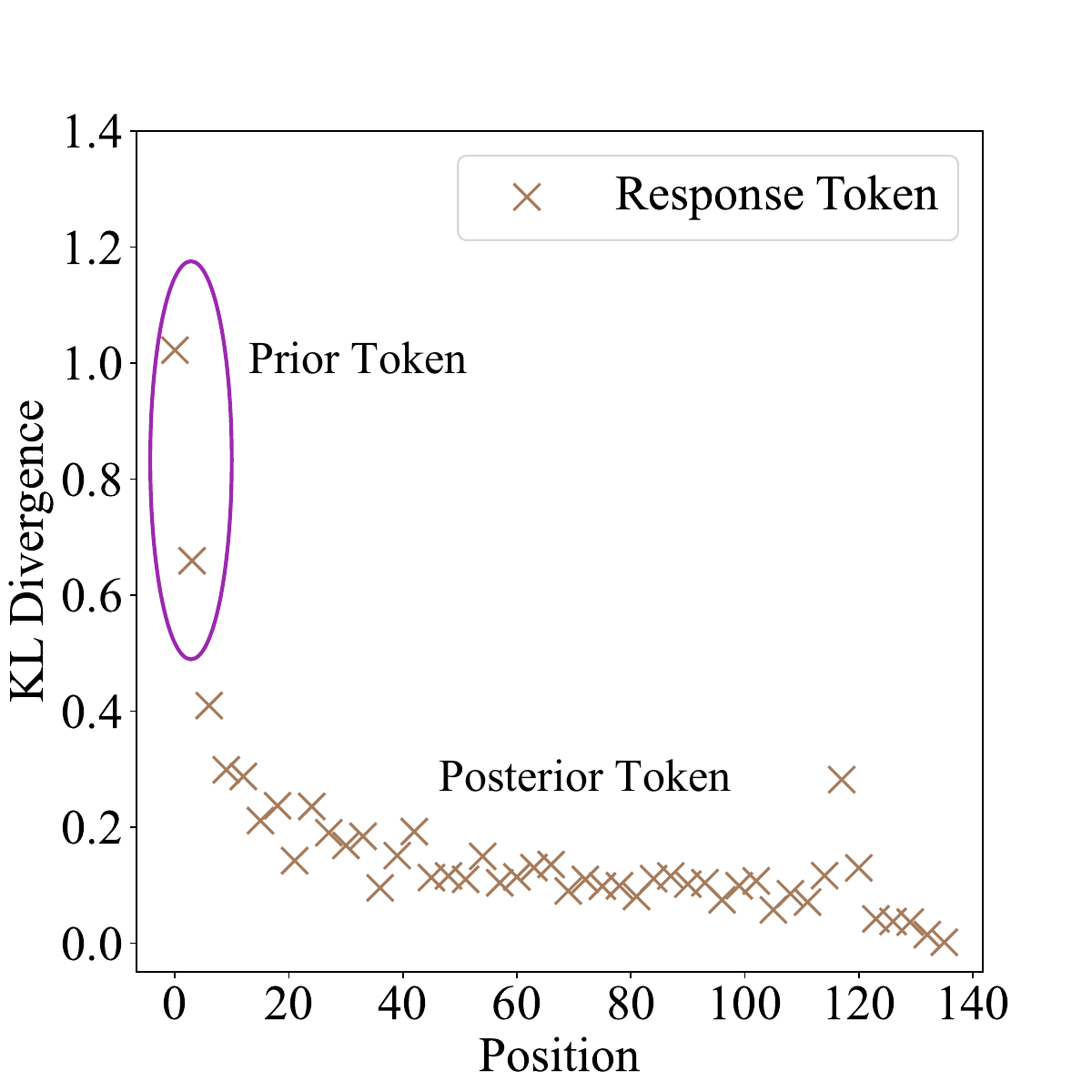}
    \label{subfig:llama_output_exp}
    \caption{Output Experimental Group}
  \end{subfigure}
  \begin{subfigure}[b]{0.24\linewidth}
    \includegraphics[width=\linewidth]{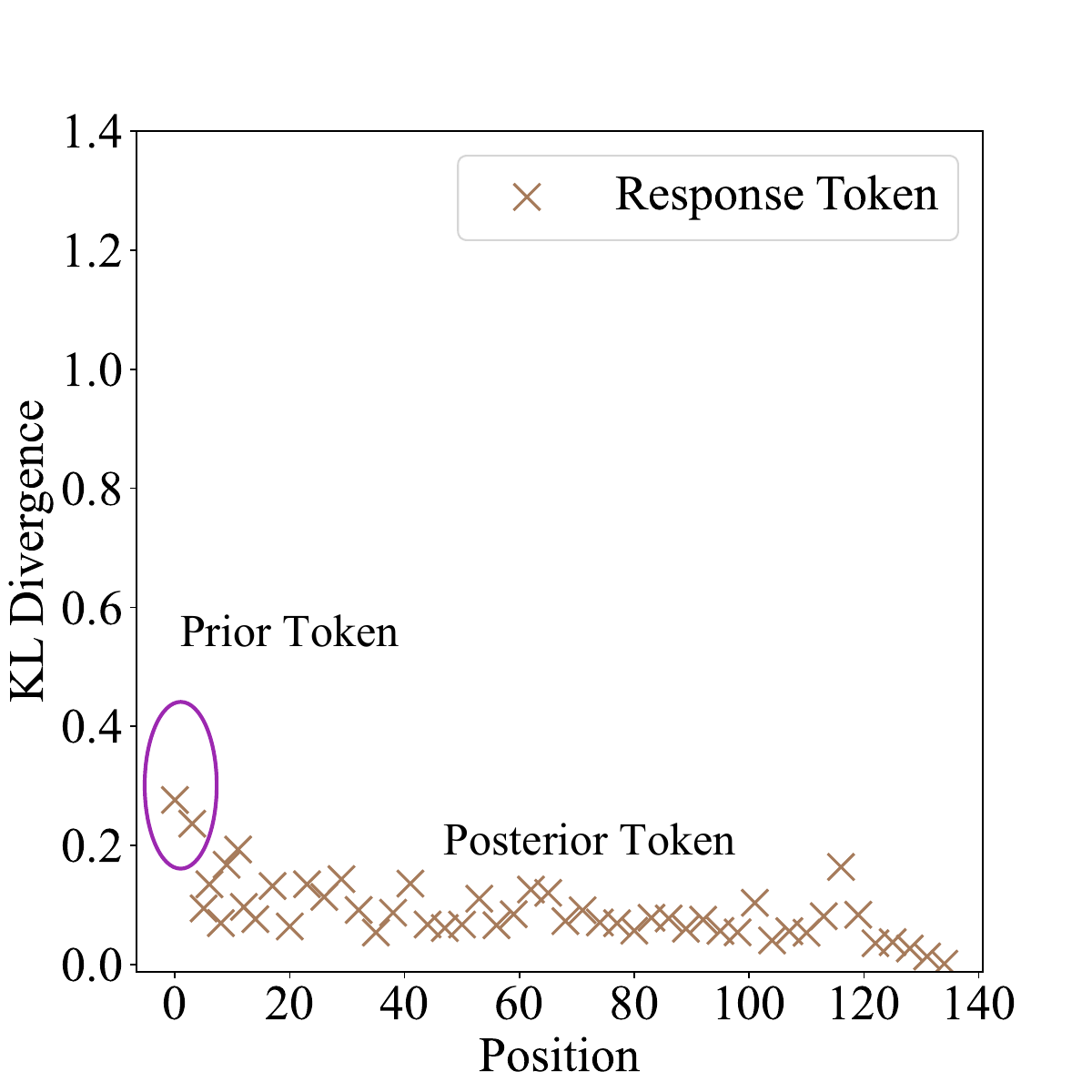}    
    \label{subfig:llama_output_con}
    \caption{Output Control Group}
  \end{subfigure} 

  \caption{The KL-divergence of token probability distributions on Llama2-7b. \textit{Experimental Group} compares zero-shot and few-shot settings, while \textit{Control Group} compares two few-shot settings with different demonstrations. We visualize the input and output separately and mark the prior query tokens and prior response tokens with purple circles.}
  \vspace{-4mm}
  \label{fig:kl_score}
\end{figure*}

In this paper, we investigate the impact of demonstrations during in-context alignment. We visualize the token distribution KL-divergence of instructions and responses in zero-shot and few-shot settings (\autoref{fig:kl_score}). To reduce context noise, we set up two few-shot settings with different demonstrations as control groups and have the following observations through comparative experiments:
(1) The model likely stores the task function learned from the demonstration in the separator token representation. (2) Demonstrations play a crucial role in prior response generation but are redundant in posterior response generation.
These observations highlight the influence of demonstrations on token representation in ICL for alignment tasks, indicating that demonstrations are not always indispensable during the entire response generation stage.

Motivated by these findings, we propose a \underline{P}rogressive \underline{I}n-\underline{C}ontext \underline{A}lignment (\textsc{\textbf{PICA}}) method to enhance both the efficiency and effectiveness of regular ICL. Specifically, Our approach involves a two-stage progressive generation strategy: the few-shot stage and the zero-shot stage. During the few-shot stage, the model generates prior part of the response using the standard ICL settings. After generating a specific number of tokens, we transition the model into the zero-shot stage, eliminating the need for further demonstrations to generate the remaining part of the response.
To capitalize on the task-related information embedded in the separator tokens, we introduce an ICL vector guidance method. Inspired by the work of task vector in ICL~\cite{icl_vector,function_vector,state_vector}, we extract the ICL vector from the hidden states of specific transformer layers. This vector is then used to steer the model during the zero-shot stage by intervening in the forward pass.
\textsc{PICA} minimize the need for demonstrations while improving output quality, thereby reducing the computational cost associated with demonstrations and enhancing overall performance.
Extensive experiments show that \textsc{PICA} outperforms regular ICL in both of efficiency and effectiveness. As a training-free method, it is also comparable to other alignment methods (i.e., SFT and RLHF). For example, on average, our PICA boosts the performance of Mistral-7b to reach 90\% of the performance of GPT-4-0613. These results support our observations and show the effectiveness of our method in various aspects of alignment. Additionally, we conduct ablation studies to investigate the robustness and generalizability of our method.
Our contributions are summarized as follows:
\begin{itemize}[leftmargin=*]
\item We delve into the impact of demonstrations on token representation in ICL and qualitatively explore the working mechanism of task functions learned from demonstrations in complex alignment tasks.
\item We propose a progressive in-context alignment method that incorporates progressive generation and ICL vector guidance. This method efficiently aligns models and significantly reduces the computational cost associated with demonstrations.
\item We conduct extensive evaluation and ablation experiments on the proposed method, where the results have fully demonstrated its efficiency and effectiveness. Our experiments and analyses provide in-depth insights for future research on in-context alignment.
\end{itemize}

\section{Related Work}

\paragraph{LLM Alignment.} Prior works have explored alignment tuning through supervised fine-tuning using public instruction datasets~\citep{DBLP:conf/emnlp/WangMAKMNADASPK22,DBLP:conf/nips/ZhouLX0SMMEYYZG23,DBLP:journals/corr/abs-2009-01325} or reinforcement learning from human feedback~\citep{DBLP:journals/corr/abs-2009-01325,DBLP:conf/nips/RafailovSMMEF23}. A common approach is to fine-tune models using instruction data to enable them to follow instructions effectively. To rapidly accumulate a vast amount of instruction tuning data, \citet{DBLP:conf/nips/WangIDHKCWMSBH23} proposes a pipeline to obtain instruction data from powerful models, such as GPT-4. LIMA leverages only 1000 high-quality instruction data points to fine-tune a 65B parameter LLM~\citep{DBLP:conf/nips/ZhouLX0SMMEYYZG23}. It shows that the minimal tuning surprisingly results in a high win rate against ChatGPT. 
Following instruction fine-tuning, the reinforcement learning is applied to further align the models~\citep{DBLP:journals/corr/abs-2009-01325}. \citet{DBLP:conf/nips/RafailovSMMEF23} introduces a training method for alignment that does not require a reward model. Its powerful convenience and effectiveness have made it one of the \emph{de facto} methods.
However, these methods necessitate substantial resources and there is evidence to suggest that such training approaches cause model forgetting of previously acquired knowledge in base LLMs~\citep{DBLP:conf/nips/WangIDHKCWMSBH23,DBLP:journals/corr/abs-2309-16155,DBLP:conf/emnlp/WangMAKMNADASPK22}.
In contrast to training-based methods, \citet{unlock_spell} experiment with ICL for LLM alignment and Confirm the feasibility of ICL for the alignment task. 
Building on this finding, we explore a training-free ICL approach. We do not merely utilize ICL. Instead, we initially investigate its working mechanism in token representation learning. This investigation helps enhance the effectiveness of in-context alignment.
Similar to us, a very recent concurrent work~\citep{DBLP:journals/corr/abs-2404-16766} also identifies the critical role of prior answer token selection in alignment tasks, and proposes a SFT model or external resources guided generation method for multilingual instruction following. Differing from their approach, we  focus on the working mechanisms and optimization methods of ICL in the mainstream English alignment tasks.

\paragraph{In-context Learning Working Mechanism.} 

Recent studies have explored the working mechanisms within ICL. Several works try to theoretically demonstrate a strong similarity between the attention patterns in ICL and the process of gradient descent~\citep{DBLP:conf/iclr/AkyurekSA0Z23,DBLP:conf/acl/DaiS0HMSW23}.  From a more practical perspective, another line of research suggests that the ICL may function by learning a mapping function from demonstrations, which it then applies to input queries to make predictions~\citep{icl_vector,function_vector,state_vector}. \citet{icl_vector}  extract an ICL task vector from the hidden states and utilize it for intervention during zero-shot inference. \citet{function_vector} extract a function vector from attention activations using the causal mediation method, which is subsequently added to the hidden states of certain transformer layers during inference. \citet{state_vector} derive a state vector from attention activations and propose several optimization strategies. Unlike these works, we focus on using comparative experiments to explore the impact of demonstrations on token representation, and leverage these findings to enhance the efficiency of ICL.

\section{Motivation}
\label{sec:motivation}


In this section, we aim to shed light on the working mechanisms of in-context learning by investigating the following question: \textbf{What is the impact of demonstration on token representation in in-context alignment?} To explore this, we design a comparative experiment to highlight how token representations differ between zero-shot and few-shot settings. We use token probability distributions as a proxy for token representations and utilize KL-divergence to measure the shifts in these distributions. By visualizing and quantifying the shifts in token probability distributions caused by demonstrations, we can understand the role of demonstrations in aligning the model and provide further optimization for in-context alignment.

Regarding the experimental setup, we randomly selected 100 data instances of similar length from Ultra-chat~\citep{DBLP:conf/emnlp/DingCXQHL0Z23}, a commonly used dataset for alignment tuning, as our experimental dataset. For the input prompt, we use a straightforward design by adding several tokens at the end of the query to serve as separator tokens, explicitly distinguishing between the query and the response. We present the visualization results based on the Llama2-7b model in the~\autoref{fig:kl_score}, while the results for other models are provided in Appendix~\ref{sec:Motivation_appendix}.
We break the token distribution of the whole instance into the input and output parts. A straightforward reason is that the input token distribution shift represents differences in understanding the instruction, while the output token distribution shift represents the ability to respond.
By observing and analyzing the visualization, we have two hypotheses: (1) the ICL alignment task function might be encoded into the separator token representation. (2) the quality of response is highly reliant on the quality of prior response tokens.


\paragraph{Input Token Distribution.} By comparing the input token probability distributions between zero-shot and few-shot settings, a significant shift is observed in both the prior tokens of the query and the separator tokens. The KL-divergence decreases as the number of query tokens increases. By comparing the experimental group and the control group, we find that the shift in the query distribution also occurs in the control group. However, this shift in the separator tokens is not consistent across different demonstration settings, suggesting distinct underlying causes for these shifts. We attribute the shift in the query’s prior token distribution to a ``context shift'', and we attribute the shift in the separator tokens distribution to a ``task shift''.
Given that LLMs are trained to predict the next token based on the provided context, altering the context directly impacts the token distribution, which we refer to as the ``context shift''. However, as the number of query tokens increases, the decision space gradually aligns for both zero-shot and few-shot settings, leading to higher consistency in query token prediction and thus a reduced KL-divergence.
On the contrary, the trend observed in the query distribution is not mirrored in the separator token distribution. In the control group, the separator token representations remain highly similar. We attribute the large KL-divergence observed in the separator token distribution of the experimental group to the differing tasks, indicating that separator tokens likely encode task-specific information during ICL. We reasonably speculate that the primary impact of demonstration on instruction understanding is reflected in the encoding of separator tokens, where the alignment task function learned through ICL is stored. This hypothesis aligns with prior work~\cite{icl_vector,state_vector}, yet our findings contribute additional evidence supporting this perspective.

\paragraph{Output Token Distribution.} Observing the visualization of output token distribution, we find that when comparing zero-shot and few-shot settings, the response token distribution shows similarity in the posterior tokens. This indicates that the model selects posterior tokens with high consistency in both zero-shot and few-shot settings. 
When comparing the prior response tokens of the experimental group and the control group, we observe a pattern similar to that of the separator tokens, suggesting that demonstrations play a crucial role in the prior response tokens.
Based on these observations and analyses, we speculate that the primary impact of demonstrations on response generation is reflected in the generation of prior answer tokens. Compared to zero-shot settings, demonstrations guide the generation of accurate prior response tokens, which implicitly helps the model successfully follow the instructions. This observation also suggests that once the prior response tokens are determined, the influence of the demonstration diminishes and becomes redundant.

\section{Method}

\begin{figure}[!t]
    \centering
    \includegraphics[width=\linewidth]{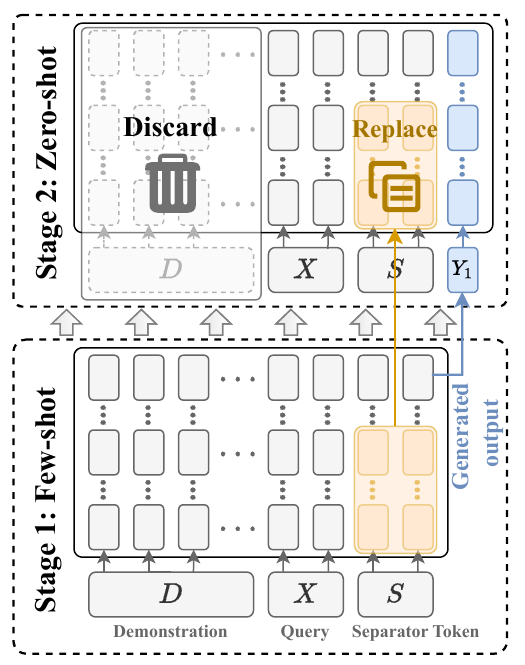}
    \caption{Overview of \textsc{PICA}, which include \textit{few-shot} stage and \textit{zero-shot} stage. The \textcolor[RGB]{96, 96, 96}{gray block} denotes the hidden state and \textcolor[RGB]{163,118,0}{orange block} denotes the separator token hidden state that forms the ICL vector. The \textcolor[RGB]{108,142,191}{blue block} denotes the generated answer token from few-shot stage.}
    \vspace{-4mm}
    \label{fig:overview}
\end{figure}

Observations from \S\ref{sec:motivation} reveal that demonstrations are not always indispensable during the entire response generation stage. To minimize the need for demonstrations while preserving alignment performance, we introduce a progressive in-context alignment approach. This methodology enhances the efficiency and efficacy of in-context alignment through two innovations: (1) a progressive generation strategy that reduces the computational cost associated with demonstrations, and (2) in-context learning vector guidance that compresses the task function from demonstrations to assist in high-quality response generation.

Inspired by underscoring the redundancy of demonstrations once the pivotal prior response tokens are determined, we introduce a progressive generation strategy, dividing response generation into few-shot and zero-shot stages.
During the few-shot stage, the model generates a specific number of prior response tokens by employing a standard in-context learning:
\begin{equation}
    Y^{\text{few}}_{i} = \mathop{\arg\max}_{Y \in V} P(Y|D,Q,S,Y^{\text{few}}_{1:i-1}),
\end{equation}
where $D$ is the demonstration, $Q$ is the query, $S$ is the separator token, and $Y^{\text{few}}_i$ is the $i$-th answer token generated in few-shot stage.
After obtaining several prior answer tokens, the model operates within a more certain and simplified decision space for token generation, allowing the omission of the demonstration to reduce computational costs. Therefore, in the zero-shot stage, the model completes the response based on the existing prior response tokens:
\begin{equation}
    Y^{\text{zero}}_{i} = \mathop{\arg\max}_{Y \in V} P(Y|Q,S,Y^{\text{few}}_{1:N},Y^{\text{zero}}_{1:i-1}),
\end{equation}
where $N$ is the number of prior tokens, and $Y^{\text{zero}}_i$ the $i$-th answer token generated in zero-shot stage.

\paragraph{In-context Learning Vector Guidance.} We observe that transformers show task-specific encoding behavior with the separator token. Recent works~\cite{icl_vector,function_vector} have similar observations, demonstrating that functions learned by ICL can be represented through compressed vectors derived from transformers and can perform simple generation tasks in zero-shot settings.
Building on this, we propose the ICL vector guidance to assist the model in generating high-quality responses during the zero-shot stage.
Unlike these previous works that intervene single hidden state of the last separator token, we intervene in the initial $L$ layer of all separator tokens. Our preliminary experiments found that this method is more effective for the alignment task, where the output is much longer than that of the simple generation tasks focused on in previous works.

Specifically, during the forward pass in the few-shot generation, we extract the separator token hidden state $H^{\text{few}}_{i}$ from the first $L$ layers, which we combine and refer to as the ICL vector. 
Subsequently, in the zero-shot stage, we intervene in the separator token representation by replacing the hidden state with the extracted hidden state from the few-shot stage:
\begin{equation}
    H^{\text{zero}}_{i} =  \
    \begin{cases} 
    H^{\text{few}}_{i} & \text{if } i \leq L \\
    \text{Layer}(H^{\text{zero}}_{i-1}) & \text{otherwise}
    \end{cases},
\end{equation}
where $\texttt{Layer($\cdot$)}$ is the process function of transformer layer. By intervening with the ICL vector, the model receives implicit guidance from the demonstration during generation, thereby improving the quality of the zero-shot stage responses.

Overall, Our progressive in-context alignment process is: In the few-shot stage, we utilize standard ICL to generate pivotal prior response tokens while extracting the ICL vector from the separator token representation.
Subsequently, we discard the demonstration and employ the ICL vector to guide the model in generating the complete response in the zero-shot setting.
This dual-stage progressive in-context alignment approach fully capitalizes on the potential of the ICL vector and the text completion capabilities of foundational language models in the zero-shot setting. By effectively harnessing these capabilities, the approach not only reduces computational cost but also maintains high fidelity in response generation across various settings.

\section{Experiment}

\subsection{Datasets and Models}

Recent research demonstrates that utilizing powerful AI assistants such as ChatGPT and GPT-4 for scoring and comparing  achieves close alignment with human evaluations while reducing costs~\citep{DBLP:conf/emnlp/LiuIXWXZ23,DBLP:journals/corr/abs-2404-04475}. Consequently, we evaluate our method using two automatic alignment benchmarks: alpaca-eval (2.0)~\citep{DBLP:journals/corr/abs-2404-04475} and just-eval~\citep{unlock_spell}.
Alpaca-eval comprises 805 instructions and provides a length-controlled win rate from the judge model by comparing the assessed results with those from a reference model. For fast and validated evaluation, we select GPT-3-text-davinci-003 and GPT-4 as reference models, while employing GPT-4-0314 as the judge model.
Just-eval includes 800 regular instructions and 200 red-teaming and malicious instructions selected from diverse open-source datasets, offering detailed evaluations across six aspects. On each aspect, scores range from 1 to 5, representing the degree of evaluation. In line with prior work~\citep{unlock_spell}, we use GPT-4-0314 as the evaluator and report the performance across three random seeds. 
For efficiency analysis, we evaluate the average inference time on 1000 test data with strictly generated 4096 tokens without using any additional decoding optimization techniques. We report the speedup compared to the standard ICL.

We conduct our experiments using three principal fundamental LLMs: Llama2-7b, Llama2-13b~\citep{DBLP:journals/corr/abs-2307-09288} and Mistral-7b (v0.1)~\citep{DBLP:journals/corr/abs-2310-06825}. These models are selected based on their moderate sizes, open-source availability, and proficiency in ICL. For comparative analysis, we utilized their respective alignment-tuned versions: Llama2-7b-chat, Llama2-13b-chat, and Mistral-7b-Instruct, facilitating a direct comparison with SFT and RLHF. Additionally, our study includes results from OpenAI's GPT models (i.e., GPT-3.5-turbo-0611 and GPT-4-0613), allowing comparison with the state-of-the-art AI assistants.  We follow the inference guidelines provided by the authors of these tuned models.

\subsection{Implementation Detail}

For the in-context learning prompt, we follow previous work~\citep{unlock_spell} and use the mainstream system message employed in aligned LLMs. We meticulously designed the demonstrations for in-context learning, creating six examples for alpaca-eval and three examples for just-eval, as they emphasize different evaluation aspects. We utilize greedy generation with a beam size of 1 and set the maximum token length to 4096.
The in-context learning vector guidance method we described earlier has a key hyper-parameter, specifically the layer $L$. Previous studies~\citep{icl_vector} have demonstrated that the choice of $L$ influences performance. We determine the intervention layer based on the win rate on alpaca-eval. We set the number of prior tokens to 10 as a trade-off between generation quality and efficiency. For consistency and reproducibility, we apply greedy decoding across all experiments. 
All experiments were conducted on a single NVIDIA A800 80G GPU, with each experiment consuming between 3 to 5 hours of GPU time, depending on the dataset and models used.

\subsection{Baseline}
\label{sec:Baseline}
In the paper, we compare our method with the following methods and ablation variants:
\begin{itemize}[leftmargin=*]
    \item \textbf{SFT or RLHF} is the baseline with alignment tuning method. We strictly follow the guidelines provided by the creators of these tuned models during inference.
    \item \textbf{Zero-shot} is the baseline for the zero-shot setting that uses only the given query as input, and \textbf{Vanilla ICL} is the regular ICL which makes predictions on the label by taking both demonstration and instruction.
    \item \textbf{Vec.} is the ablation variants that only utilize ICL vector guidance in zero-shot setting, while \textbf{Prog.} is the ablation variants that apply progressive generation strategy without ICL vector guidance during zero-shot stage. 
\end{itemize}

\subsection{Main result}


\begin{table*}[ht!]
\centering
\small
{
\setlength{\tabcolsep}{4pt}
	\begin{tabular}{l|cc|ccc ccc|c}
		\toprule
        \multirow{2}{*}[-2.5pt]{\textbf{Models + Alignment Methods}}       &\multicolumn{2}{c|}{\textbf{Alpaca-eval}}    &\multicolumn{6}{c|}{\textbf{Just-eval}} &\multirow{2}{*}[-2.5pt]{\textbf{Speedup}}  \\ 
        \cline{2-9}
		                          &\multicolumn{1}{c}{vs GPT-3} &\multicolumn{1}{c|}{vs GPT-4}   &\multicolumn{1}{c}{Helpful}    &\multicolumn{1}{c}{Clear}      &\multicolumn{1}{c}{Factual}    &\multicolumn{1}{c}{Deep}   &\multicolumn{1}{c}{Engaging}   &\multicolumn{1}{c|}{Safe}  \\
        \hline
        GPT-3.5-turbo-0611              &69.51   &46.46     &4.82   &4.97   &4.84   &4.33   &4.66   &4.99   &-  \\
        GPT-4-0613                      &72.51   &53.52     &4.86   &4.99   &4.90   &4.49   &4.61   &4.97   &-  \\
        \hline\hline       
        Llama2-7b-chat (RLHF)           &40.50          &17.49          &4.12           &\textbf{4.84}  &4.13           &\textbf{4.18}  &\textbf{4.77}  &\textbf{5.00}  &5.68  \\
        Llama2-7b (Zero-shot)           &24.65          &11.74          &2.78           &3.01           &3.11           &2.27           &2.29           &1.05           &5.81  \\
        Llama2-7b (Vanilla ICL)         &42.47          &15.00          &4.01           &4.10           &4.16           &3.50           &3.31           &1.98           &1.00   \\
         
        Llama2-7b (Vec.)                &36.51          &13.73          &3.68           &3.72           &3.80           &3.01           &2.94           &1.73           &5.43   \\
         
        Llama2-7b (Prog.)               &42.13          &16.23          &3.78           &3.82           &3.94           &3.26           &3.04           &1.78           &5.53   \\
         
        Llama2-7b (PICA)          &\textbf{45.90} &\textbf{21.57} &\textbf{4.21}  &4.09           &\textbf{4.30}  &3.41           &3.42           &2.09           &5.45   \\
        \hline\hline
        Llama2-13b-chat (RLHF)          &55.30          &38.60          &4.36           &\textbf{4.94}  &4.36           &\textbf{4.55}  &\textbf{4.83}  &\textbf{5.00}  &4.97   \\
        Llama2-13b (Zero-shot)          &33.73          &15.20          &3.26           &3.65           &3.60           &2.63           &2.62           &1.86           &5.31   \\
        Llama2-13b (Vanilla ICL)        &59.82          &37.61          &4.38           &4.70           &\textbf{4.68}  &4.37           &4.24           &4.09           &1.00   \\
         
        Llama2-13b (Vec.)               &53.57          &24.43          &4.24           &4.45           &4.24           &3.85           &3.79           &2.22           &4.84   \\
         
        Llama2-13b (Prog.)              &58.14          &34.91          &4.25           &4.33           &4.35           &3.60           &3.48           &4.01           &4.78   \\
         
        Llama2-13b (PICA)   &\textbf{62.78} &\textbf{40.15} &\textbf{4.58}  &4.66           &\textbf{4.68}  &4.16           &4.15           &4.37           &4.83   \\
        \hline\hline
        Mistral-7b-instruct (SFT)       &62.78          &43.30          &4.72           &4.75           &4.30           &4.41           &4.37           &2.00           &4.95   \\
        Mistral-7b (Zero-shot)          &43.32          &22.55          &3.86           &4.14           &4.05           &3.38           &3.31           &1.61           &5.23   \\
        Mistral-7b (Vanilla ICL)        &62.03          &40.35          &4.70           &\textbf{4.87}  &\textbf{4.81}  &4.32           &4.38           &3.03           &1.00   \\
         
        Mistral-7b (Vec.)               &61.19          &37.61          &4.76           &4.81           &4.74           &4.36           &4.32           &2.48           &5.02   \\
         
        Mistral-7b (Prog.)              &62.75          &39.73          &4.76           &4.84           &4.77           &\textbf{4.42}  &\textbf{4.61}  &4.17           &4.83   \\
         
        Mistral-7b (PICA)   &\textbf{66.38} &\textbf{44.33} &\textbf{4.79}  &4.86           &4.79           &\textbf{4.42}  &4.59           &\textbf{4.34}  &4.93   \\   
        \hline
	\end{tabular}
 }
    \caption{Comparison of alignment performance and efficiency. Alpaca-eval presents the win rate against competitor models, while Just-eval presents the scores across six aspects (scores are on a scale of 1-5). Results highlighted in gray represent our methods: \textit{Vec.} denotes the ICL vector guidance and  \textit{Prog.} denotes progressive generation ablation variants. The best results in each aspect are marked in \textbf{bold}. Speedup indicates the efficiency improvement compared to vanilla ICL.}
    \label{tab:alignment result}
    \vspace{-0.4cm}
\end{table*}

\autoref{tab:alignment result} presents the win rates of each baseline on alpaca-eval and the scores on just-eval, as well as the speedup for efficiency analysis. In addition to our complete PICA method, we also present evaluation results for two ablation variants (i.e., `Vec.' and `Prog.') to explore the effectiveness of the two proposed innovations. The combination of these innovations constitutes our PICA method.

\nosection{PICA outperforms the baseline with tuning-free baselines.} As shown in the~\autoref{tab:alignment result}, our method outperforms zero-shot and vanilla ICL baselines across three models on alpaca-eval. On the just-eval dataset, our PICA also surpasses the tuning-free baseline in the majority of aspects. Compared to regular ICL, our method effectively improves helpfulness, factuality, engagement, and safety. However, in terms of clarity and depth, our method shows a minor decline. We attribute this to the fact that our approach still has limitations in generating consistently information-rich responses, indicating that the ICL vector cannot fully encapsulate all the information provided by the demonstration.

\nosection{PICA is comparable to the alignment tuning methods.}
When compared to SFT or RLHF models, our approach demonstrates superior performance on the alpaca-eval dataset, indicating an overall advantage over SFT and RLHF methods. However, on the just-eval dataset, the results vary across different aspects. For instance, in the aspects of helpfulness and factuality, our method excels, highlighting its capability to follow instructions and generate high-quality and accurate responses. This also supports the widespread hypothesis that alignment tuning may cause models to forget some of their knowledge~\citep{DBLP:conf/nips/WangIDHKCWMSBH23,DBLP:journals/corr/abs-2309-16155}. Conversely, in terms of clarity, depth, and engagement, our method lags slightly, suggesting that SFT and RLHF have an advantage in producing high-quality response styles over ICL. In terms of safety, our method surpasses SFT but does not exceed RLHF, indicating that ICL provides relatively basic safety alignment.
On the other hand, with strong models such as Llama2-13b or Mistral-7b, the performance of our PICA can reach 90\% of the performance of GPT-3.5 and GPT-4.

\nosection{PICA achieves high efficiency compared to vanilla ICL.} Analyzing the speedup shown in~\autoref{tab:alignment result}, our method significantly reduces the time cost compared to vanilla ICL (e.g., achieving a 5.45× speedup on Llama2-7b) and is close to the zero-shot method across three models. This improvement is attributed to our progressive generation strategy, which successfully saves a substantial amount of time by discarding the demonstration. Notably, our method is orthogonal to attention speedup techniques, such as flash attention~\citep{DBLP:conf/nips/DaoFERR22} and page attention~\citep{DBLP:conf/sosp/KwonLZ0ZY0ZS23}. We will leave further exploration for future work.

\nosection{Both progressive generation strategy and ICL vector guidance contribute to performance improvement.} We conduct ablation experiments on our proposed progressive generation strategy and ICL vector guidance, as indicated by the results highlighted in grey in ~\autoref{tab:alignment result}. When only one of these methods is used, the model's performance declines, with a more significant drop observed when the progressive generation strategy is removed. This clearly demonstrates the effectiveness of both methods, with the progressive generation strategy playing a more critical role. It also indicates the limitations of ICL vector guidance, which, while effective in simpler tasks~\cite{icl_vector,function_vector}, shows constraints in more complex alignment tasks.

Overall, our method outperforms ICL in performance and efficiency, achieving results comparable to alignment tuning. 
These promising outcomes validate the effectiveness of our approach and empirically support our understanding of the role of demonstrations in in-context alignment.

\section{Analysis}

\subsection{Layer Selection}


\begin{figure}[!t]
    \centering
    \includegraphics[width=\linewidth]{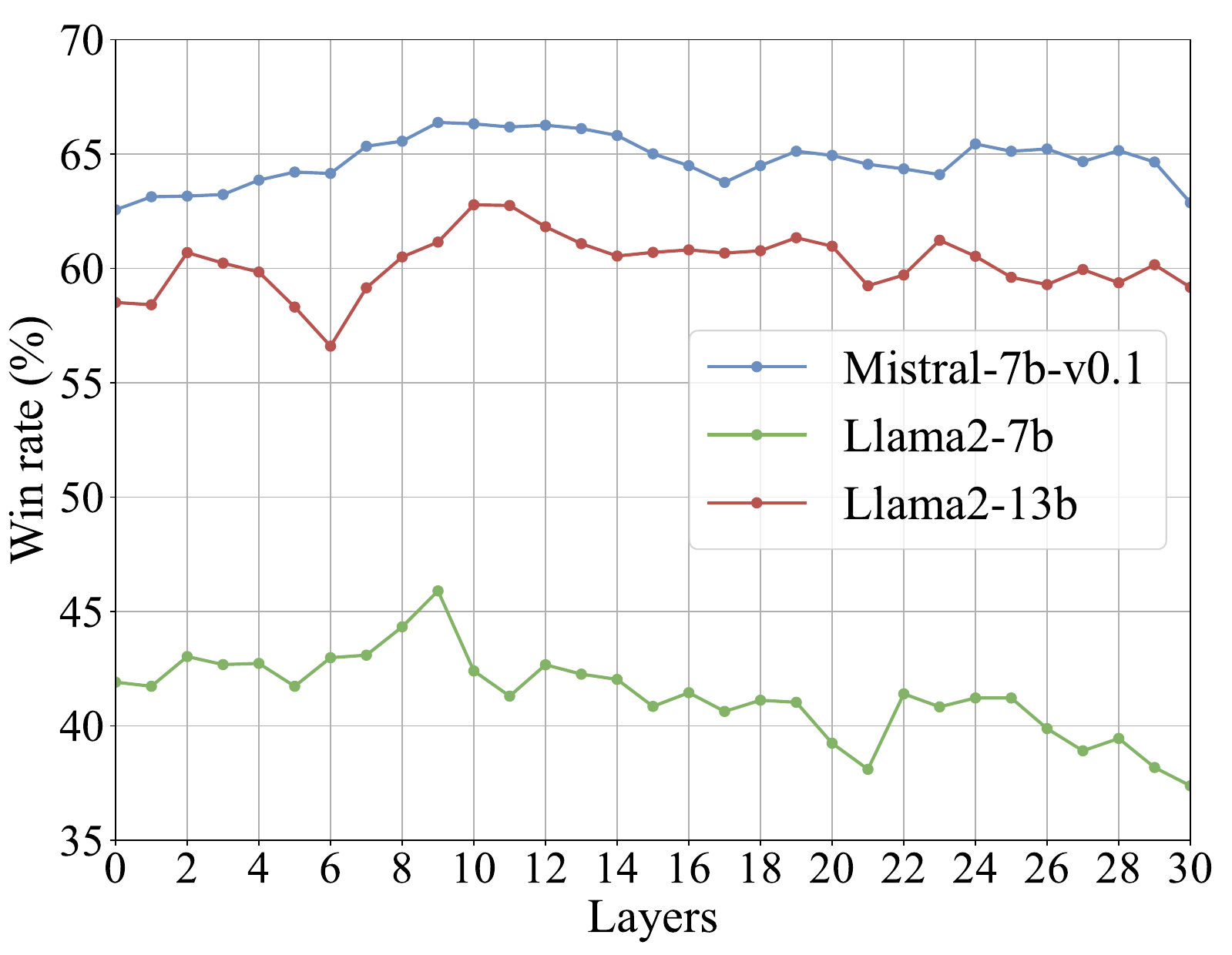}
    \caption{Win rate comparing with GPT-3-text-davinci-003 on alpaca-eval for each choice of the intermediate layer $L$.}
    \label{fig:layer_ablation}
    \vspace{-0.4cm}
\end{figure}

We delve into the impact of layer selection on the extraction of the ICL vector. We evaluate the performance based on the win rate compared to GPT-3-text-davinci-003 on the alpaca-eval datasets, as shown in~\autoref{fig:layer_ablation}.
Our results reveal a dual-phase trend: initially, increasing the number of layers improves performance, but this improvement stops or slightly declines in the later layers. This indicates that the ICL function is dynamically stored within the separator token representation. In the initial layers, transformers primarily focus on learning and encapsulating the ICL function within the hidden state, where additional layers enhance the richness of the functional information in the ICL vector. In contrast, the later layers prioritize applying this learned information for prediction tasks. Here, additional layers tend to introduce noise, causing a slight drop in performance.
This also suggests that our method is not significantly affected by layer selection, confirming the robustness of our approach.

\subsection{Prior token Ablation}

\begin{figure}[!t]
    \centering
    \includegraphics[width=\linewidth]{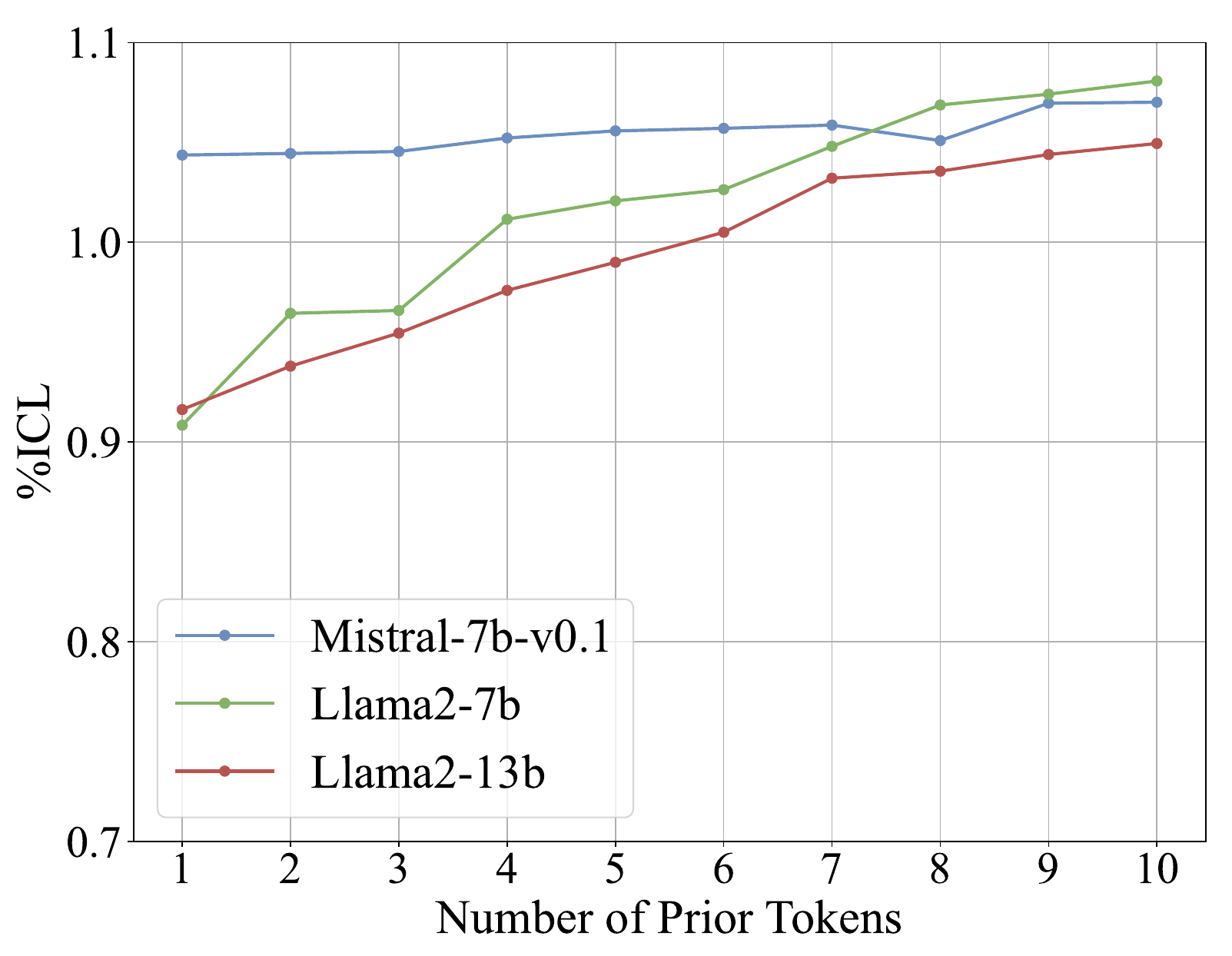}
    \caption{Win rate comparing with GPT-3-text-davinci-003 on alpaca-eval for number of the prior token on three models. We normalize the result with vanilla ICL result.}
    \label{fig:prior_ablation}
    \vspace{-0.4cm}
\end{figure}

\autoref{fig:prior_ablation} presents an ablation study on the number of prior tokens across three models, normalized by the vanilla ICL results.
An intuitive conclusion is that increasing the number of prior tokens improves the model's performance, and with about 8 prior tokens, PICA surpasses vanilla ICL. However, this improvement trend gradually diminishes. When the number of prior tokens reaches 10, the performance gain becomes less significant. This indicates that the demonstration aligns approximately the first 10 tokens to human performance. After generating 10 tokens, the base model can largely complete the response generation independently.

\subsection{Robustness Analysis}

\begin{figure}[!t]
    \centering
    \includegraphics[width=\linewidth]{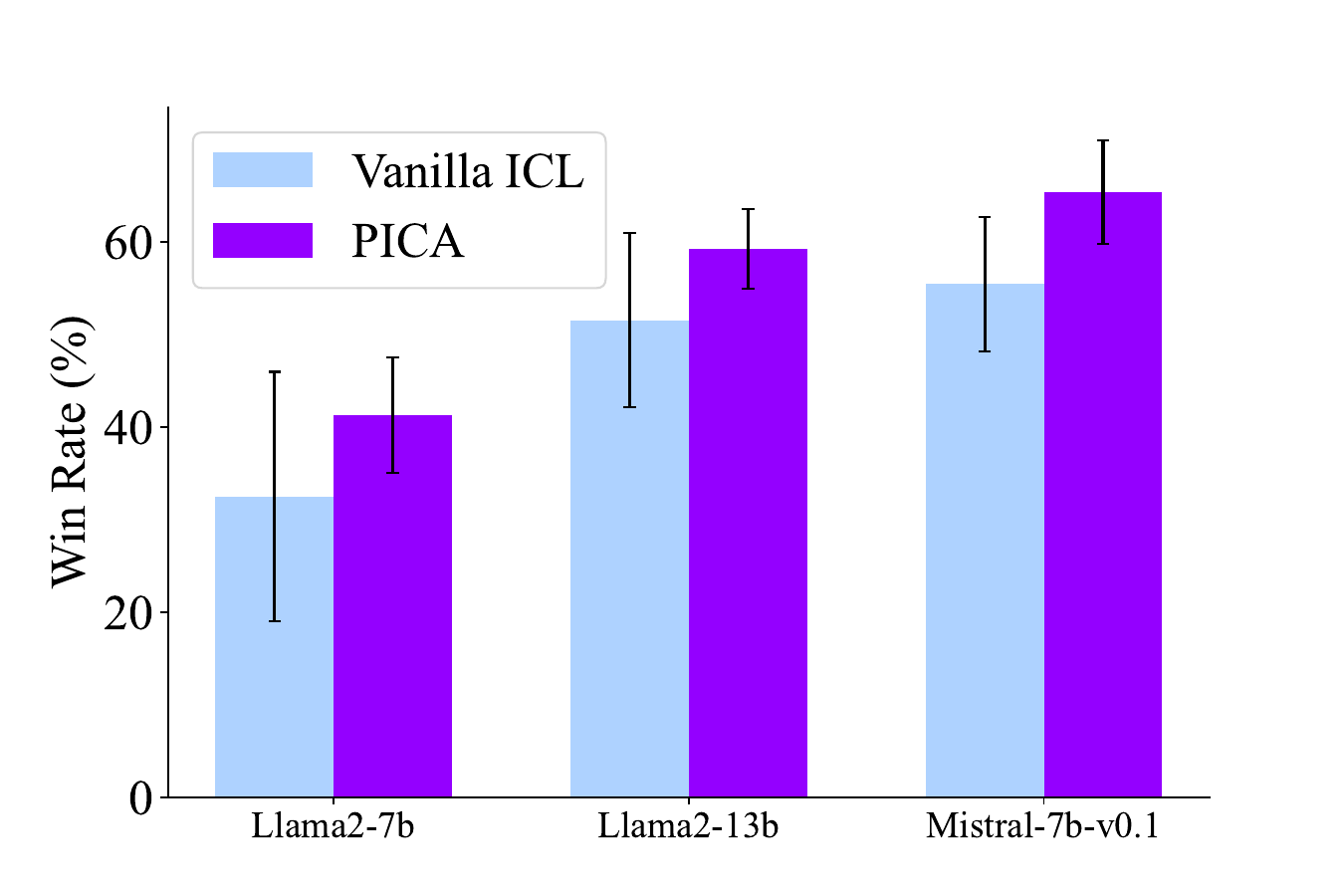}
    \caption{The mean and standard error of ICL and PICA performance with five demonstration across three models.}
    \label{fig:rubust}
    \vspace{-0.2cm}
\end{figure}

In this section, we examine the robustness of PICA to demonstration selection. Specifically, we evaluate the performance of ICL and PICA across three models using five different sets of demonstrations. The results, including the mean and standard deviation of the performance metrics, are shown in \autoref{fig:rubust}. We observe that the ICL method is more sensitive to changes in the demonstrations compared to the PICA method across all three models. This indicates that PICA effectively enhances robustness. 
We attribute this to our approach of explicitly incorporating demonstrations only in the prior response tokens, while using implicit demonstration representations during the zero-shot generation stage. This strategy effectively mitigates the impact of suboptimal demonstrations on performance.

\subsection{Human Evaluation}
We randomly sampled 100 examples each from the alpaca-eval and just-eval datasets, presenting the responses generated by PICA alongside those from the SFT or RLHF models to computer science graduate students who serve as annotators. We asked the annotators to choose which response was better or if it was a tie. \autoref{tab:humaneval} shows the results, which align with the automated evaluation.

\begin{table}[t!]
\centering
\setlength{\tabcolsep}{4pt}
	\begin{tabular}{l|c}
    \toprule
	\textbf{Winner} & \textbf{Ratio (\%)} \\
	\midrule
        Mistral-7b (\textbf{PICA})     & 35.4 \\
        Mistral-7b-instruct (SFT)               & 24.1        \\     
	    Tie & 40.5 \\
    \midrule
    Llama2-13b (\textbf{PICA})    & 34.6 \\
    Llama2--13b-chat (RLHF)                & 21.3 \\
    Tie & 44.1 \\
	\bottomrule
    \end{tabular}
    \caption{Results of human evaluation: The win rate of pairwise comparisons between PICA and SFT or RLHF.} 
    \vspace{-0.4cm}
    \label{tab:humaneval}
\end{table}

\section{Conclusion}

In this paper, we investigate and analyze the impact of demonstrations on token representation in in-context alignment through comparative experiments. Based on our observations and analyses, we introduce a novel progressive in-context alignment method that significantly reduces the need for demonstrations while preserving alignment performance. Extensive experiments indicate that PICA outperforms tuning-free baselines in both effectiveness and efficiency, achieving performance that is better or comparable to SFT or RLHF. Our experiments and analyses provide in-depth insights for future research on ICL in alignment. In the future, we aim to further explore the mechanisms and optimizations of ICL in more complex tasks.
\section*{Limitations}

Despite our discoveries and improvements, we must acknowledge certain limitations in our work:

\paragraph{Model Size:} We evaluated our method on Llama2-7b, Llama2-13b, and Mistral-7b, and these experiments were conducted on a limited scale with moderately sized models. This limits our exploration of the application of PICA on larger models. We will explore the use of PICA on larger models such as llama2-70b in future work.

\paragraph{Theoretical Foundation:} Our conclusions about the role of demonstration and ICL working mechanism lack rigorous theoretical grounding. In exploring the working mechanism of ICL, we derived some hypotheses through comparative experiments on token representation. While these hypotheses provided insights, they lack solid mathematical derivation and a theoretical basis, limiting the generalizability of our method. For example, in Appendix~\ref{sec:error}, we analyze a kind of instruction that PICA does not handle well.

\paragraph{Evaluation Datasets:} Most of our experiments utilized the alpaca-eval and just-eval datasets, which are based on AI assistant automated evaluation pipelines. Related work~\cite{DBLP:journals/corr/abs-2404-04475} has shown that these GPT-4-based evaluation methods can introduce biases, such as a preference for longer responses, which may affect the accuracy of our experimental results. Additionally, our dataset quantity is still limited, and the evaluation metrics do not fully cover all aspects of alignment, such as mathematics, reasoning, and coding. We will continue to explore our method with more comprehensive evaluation metrics in future work.

\section*{Acknowledgments}
This work is jointly supported by grants: National Natural Science Foundation of China (No. 62376067), National Natural Science Foundation of China (No. 62406088) and Guangdong Basic and Applied Basic Research Foundation (2023A1515110078).

\bibliography{custom}

\appendix

\clearpage

\begin{figure*}[!t]
    \centering
    \includegraphics[width=1.0\linewidth]{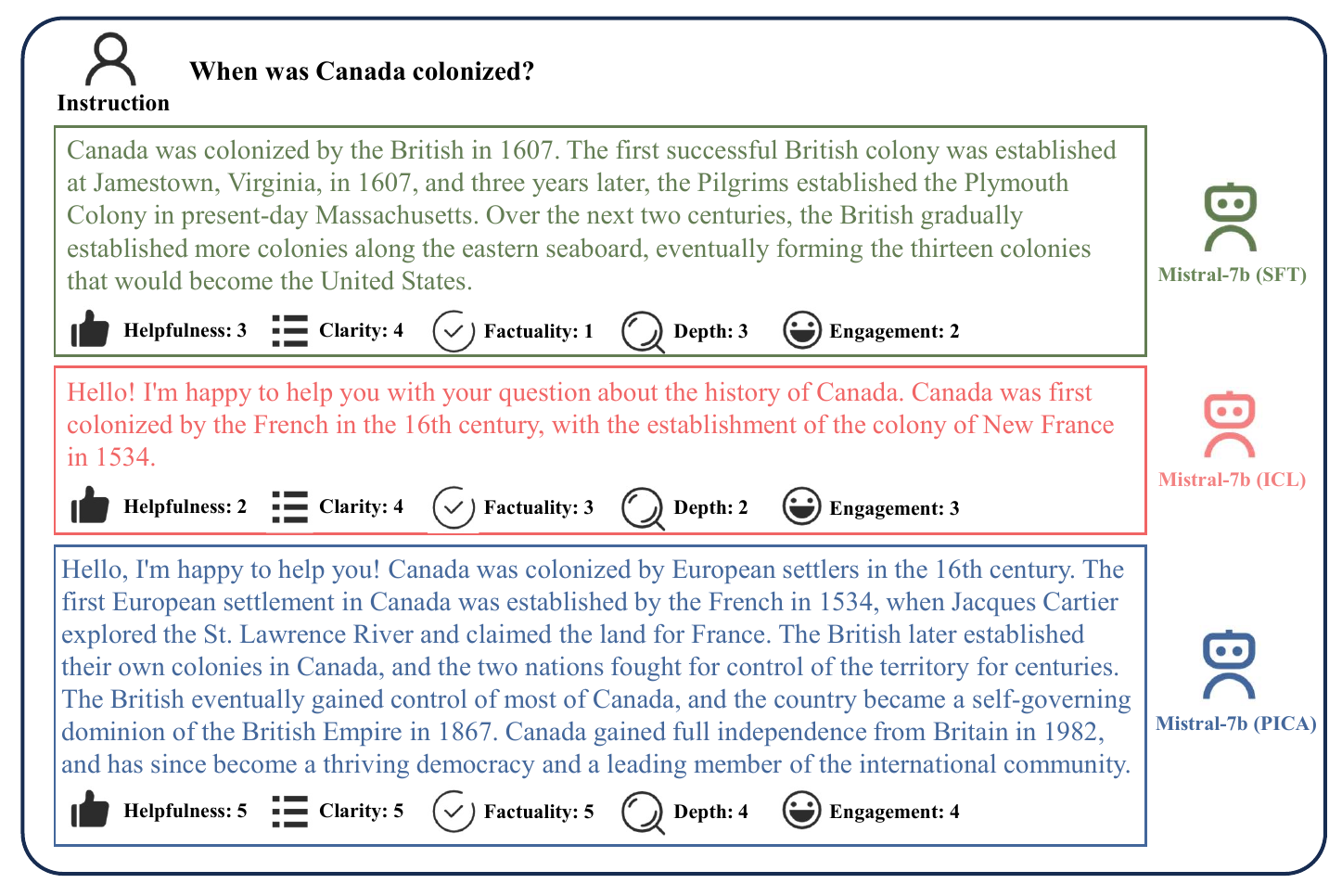}
    \caption{Case study of SFT, ICL, and PICA on Mistral-7b. We report results of the five regular evaluation aspects on just-eval.}
    \label{fig:case}
\end{figure*}

\section{Case Study}

We present a case study comparing SFT, ICL, and PICA on Mistral-7b in~\autoref{fig:case}.
The SFT model incorrectly stated that Canada was colonized by the British in 1607, leading to poor performance in factuality with a score of 1. This highlights a common issue with SFT models, where they may forget acquired knowledge over time. As a result, the SFT model received low marks in helpfulness (3) and engagement (2), despite a reasonable clarity score (4). This misrepresentation shows the limitations of the SFT approach in retaining and accurately recalling historical facts.
The ICL model is relatively better in factuality. However, the generated content lacked depth and richness, scoring 2 in depth and 2 in helpfulness, suggesting that while the ICL method generates some stylistic tokens, it does not produce sufficiently detailed or useful responses. 
Our PICA model provided a comprehensive and accurate response, detailing the colonization history of Canada, resulting in high scores across all aspects: helpfulness (5), clarity (5), factuality (5), depth (4), and engagement (4). The PICA model effectively combined stylistic tokens with detailed and accurate information, showcasing its capability to generate high-quality responses that are both informative and engaging.

\section{Error Analysis}
\label{sec:error}



\begin{figure*}[htbp!]
\centering

\begin{subfigure}[b]{0.48\linewidth}
    \includegraphics[width=\linewidth]{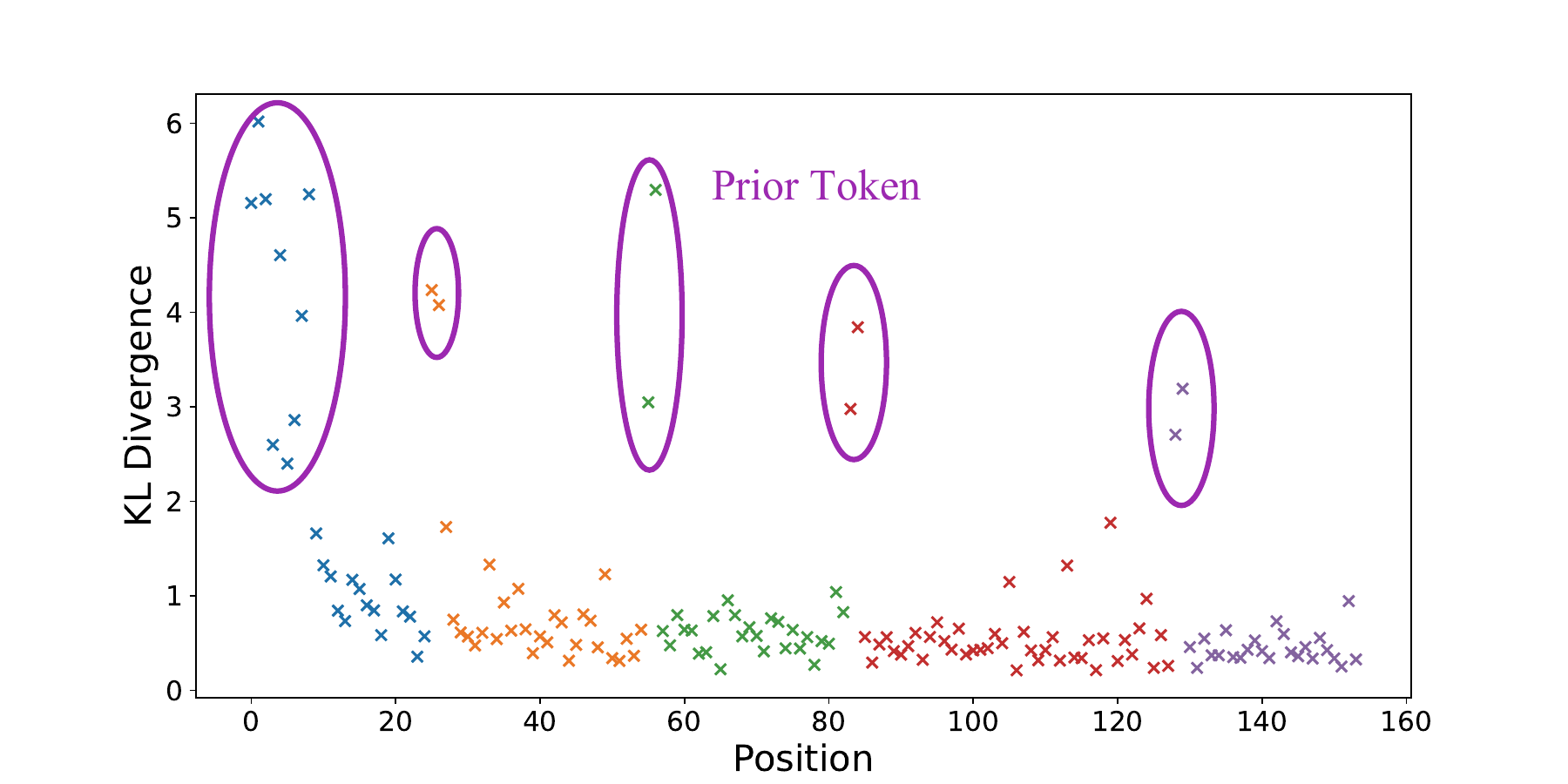}
    \caption{Llama2-7b}
  \end{subfigure}
  \begin{subfigure}[b]{0.48\linewidth}
    \includegraphics[width=\linewidth]{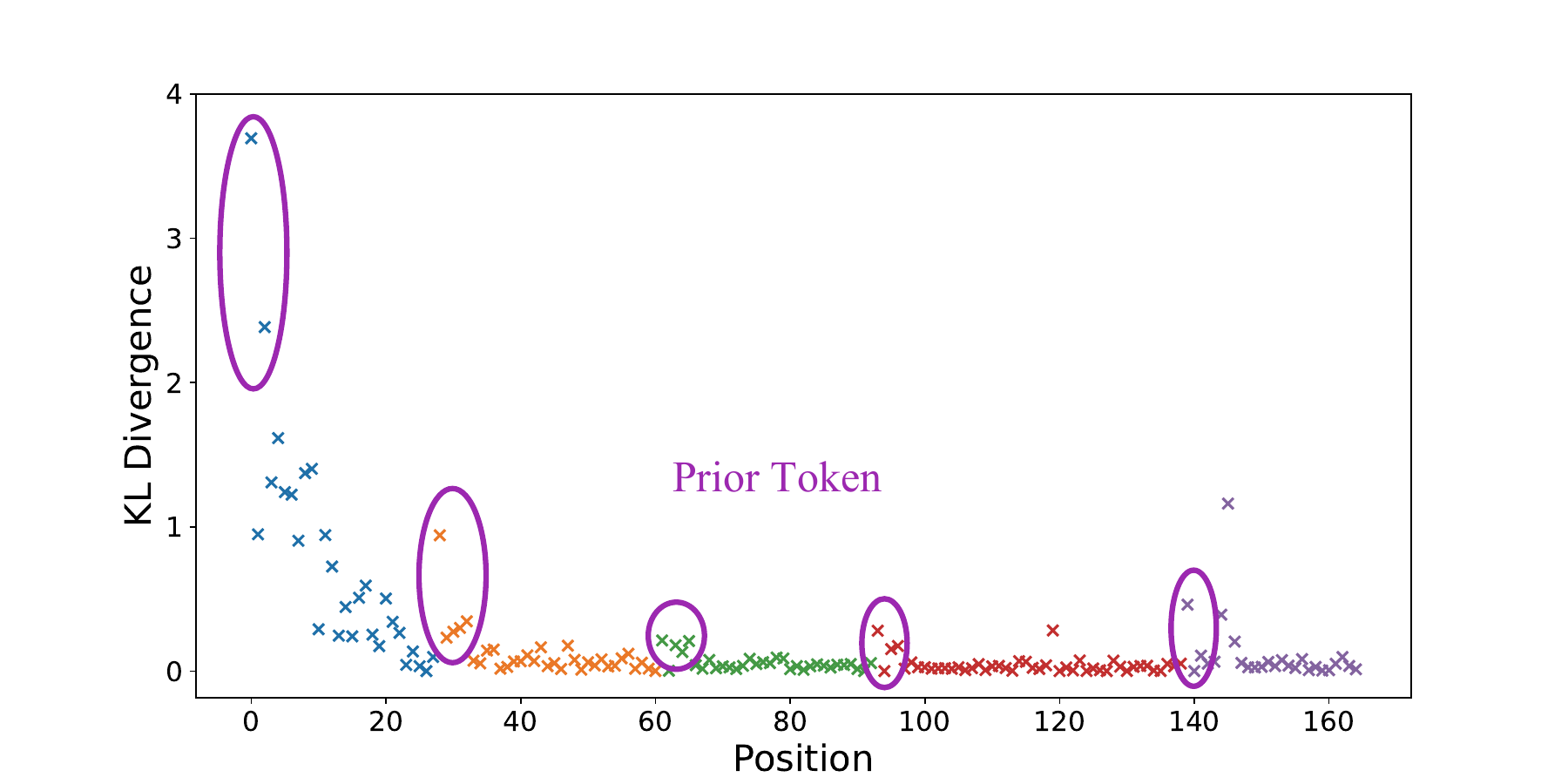}
    \caption{Mistral-7b}
  \end{subfigure}

  \caption{KL-divergence of response token distributions of enumerative instructions on Llama2-7b and Mistral-7b. }
  \label{fig:enumerate_answer}
\end{figure*}

In our preliminary experiments, we found that the proposed PICA approach frequently performed poorly in generating enumeration-type responses (e.g. ``Give me a list of some famous world music artists.''). Consequently, we analyzed the KL-divergence of responses to these instructions in zero-shot and few-shot settings. The visualization results are shown in Figure~\ref{fig:enumerate_answer}. 
Our observations indicate that, although the trend of KL-divergence is generally similar to what we observed in \S\ref{sec:motivation} there are differences in each enumeration of the response. We found that the KL-divergence of prior tokens is usually larger than the posterior tokens in each enumeration, indicating that these prior enumeration tokens are pivotal. The quality of responses to enumerative instructions is influenced not only by the selection of prior response tokens but also by the selection of prior enumeration tokens. We attribute this to the fact that each enumerated item is relatively independent of each other. When generating these enumerations, the model requires more substantial guidance from the demonstrations. However, the proposed ICL vector and the positions of previous enumeration responses do not provide enough information for generation, thus reducing the quality of each enumeration. This highlights a limitation of our current PICA approach, which we will explore and optimize in future work.

\section{More Exploration on Demonstration}
\label{sec:Motivation_appendix}

\begin{figure*}[htbp]
\centering
\begin{subfigure}[b]{0.24\linewidth}
   \includegraphics[width=\linewidth]{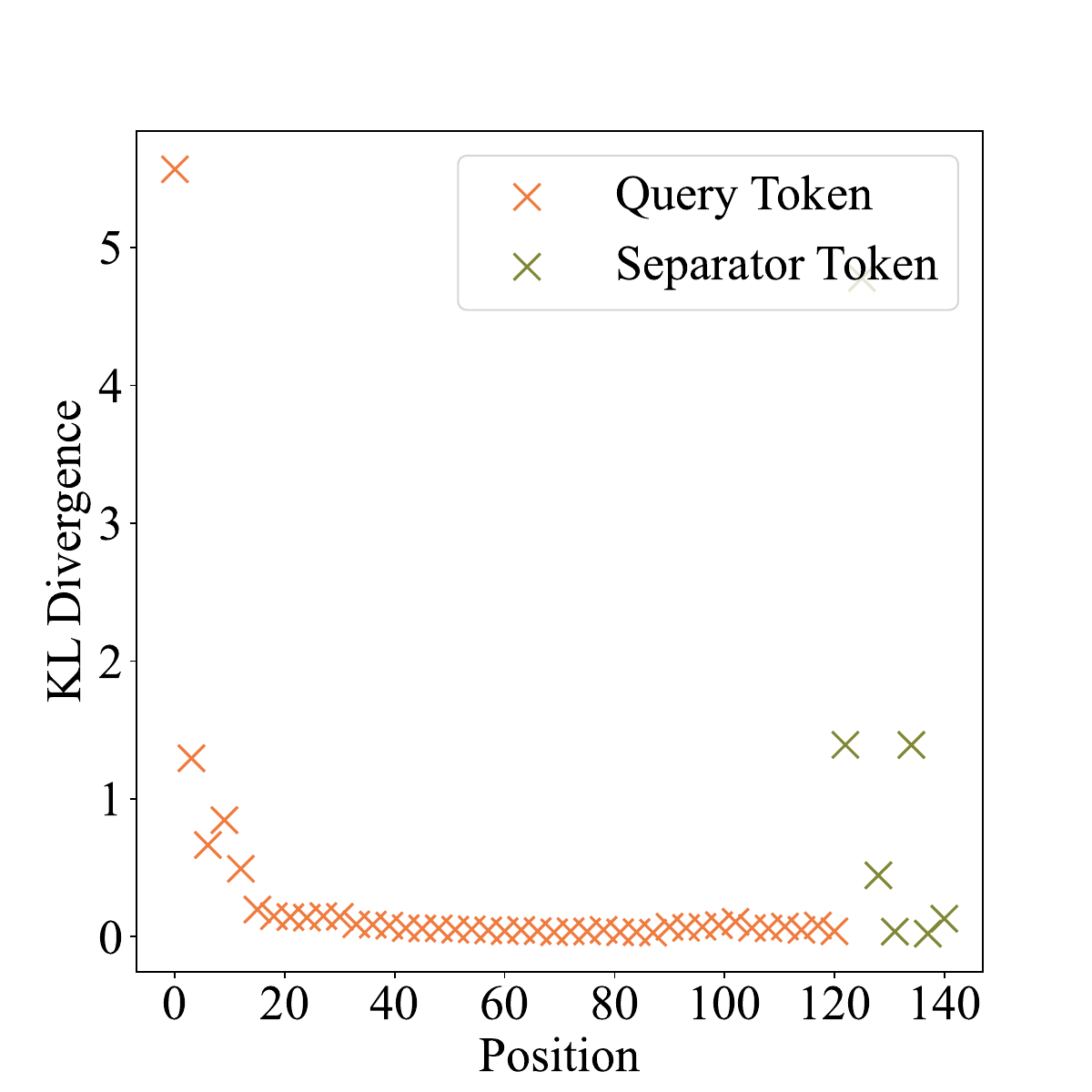}
    \caption{Input Experimental Group}
  \end{subfigure}
  \begin{subfigure}[b]{0.24\linewidth}
    \includegraphics[width=\linewidth]{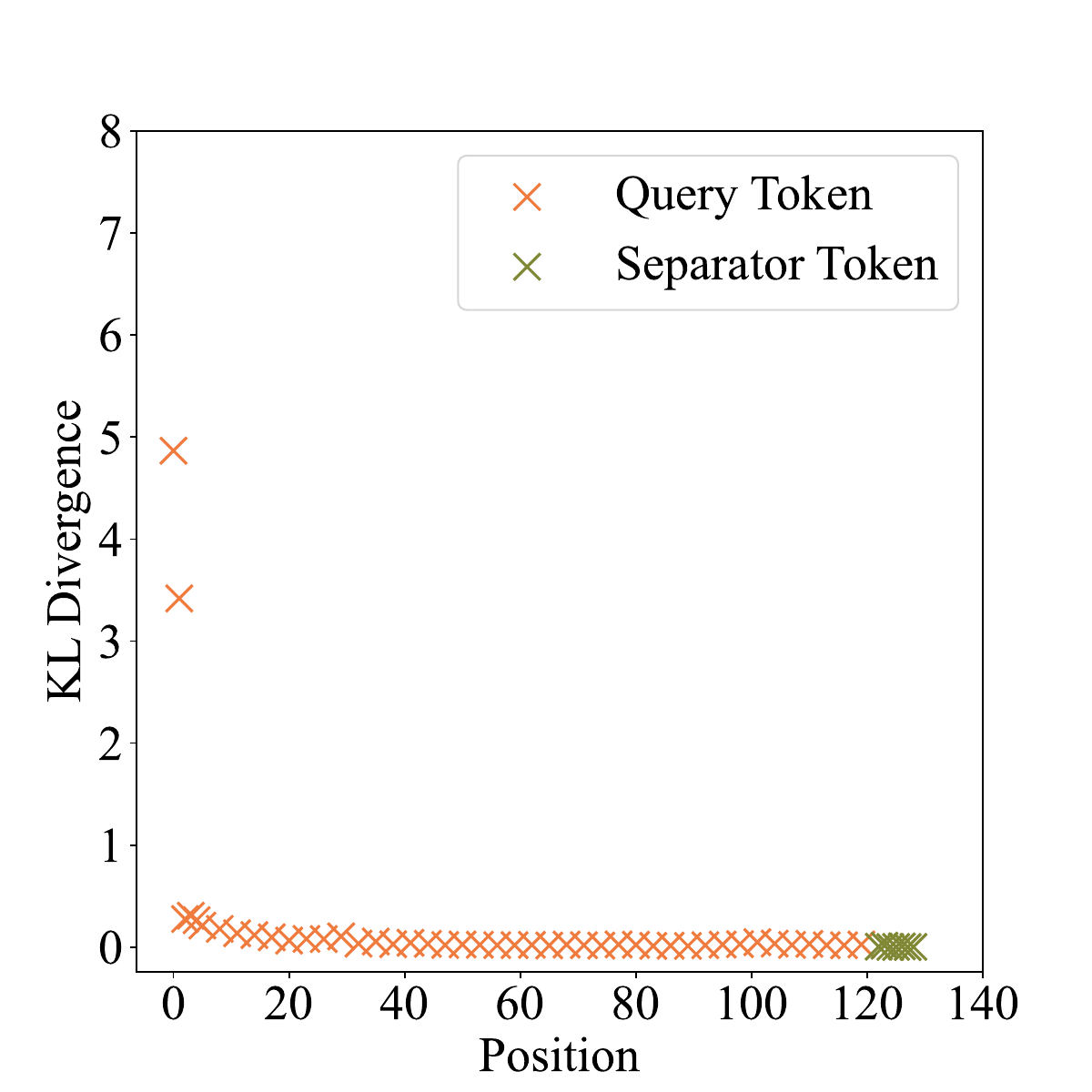}
    \caption{Input Control Group}
  \end{subfigure}
  \begin{subfigure}[b]{0.24\linewidth}
    \includegraphics[width=\linewidth]{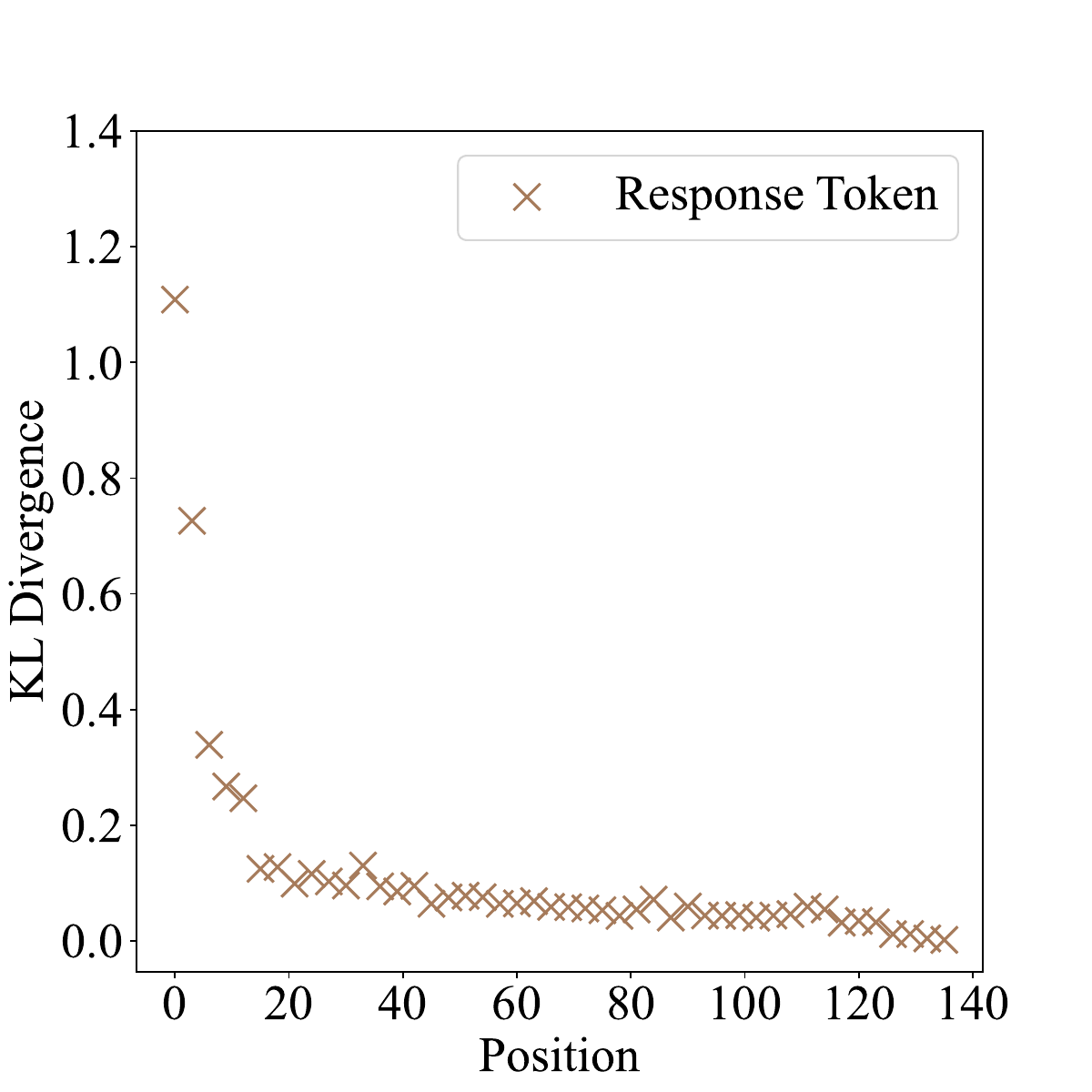}
    \caption{Output Experimental Group}
  \end{subfigure}
  \begin{subfigure}[b]{0.24\linewidth}
    \includegraphics[width=\linewidth]{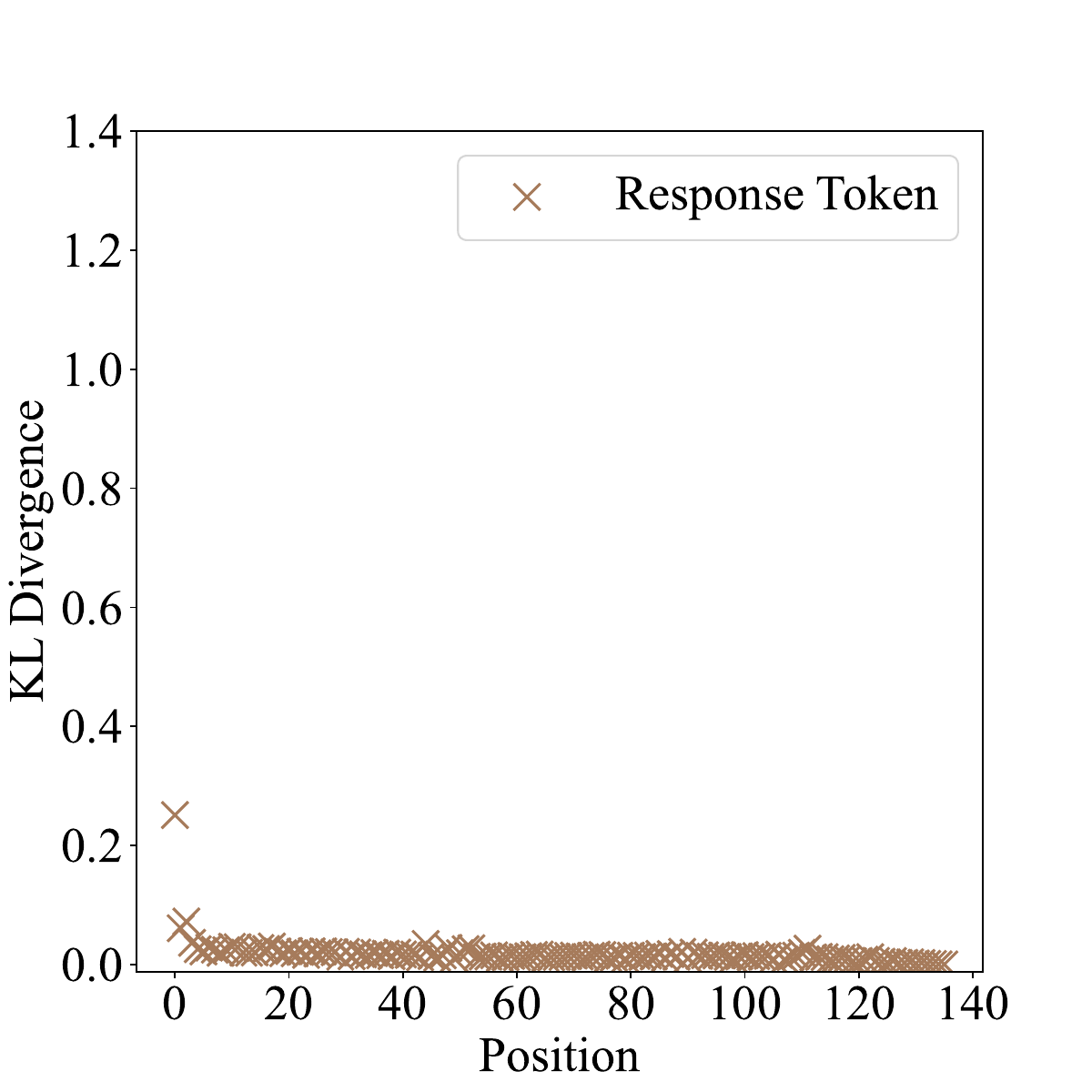}
    \caption{Output Control Group}
  \end{subfigure} 

  \caption{The KL-divergence of token probability distributions on Mistral-7b. \textit{Experimental Group} compares zero-shot and few-shot settings, while \textit{Control Group} compares two few-shot settings with different demonstrations. We visualize the input and output separately and mark the prior query tokens and prior response tokens with purple circles.}
  \label{fig:kl_score_mixtral}
  \vspace{-0.4cm}
\end{figure*}

\begin{figure*}[htbp!]
\centering
\begin{subfigure}[b]{0.24\linewidth}
    \includegraphics[width=\linewidth]{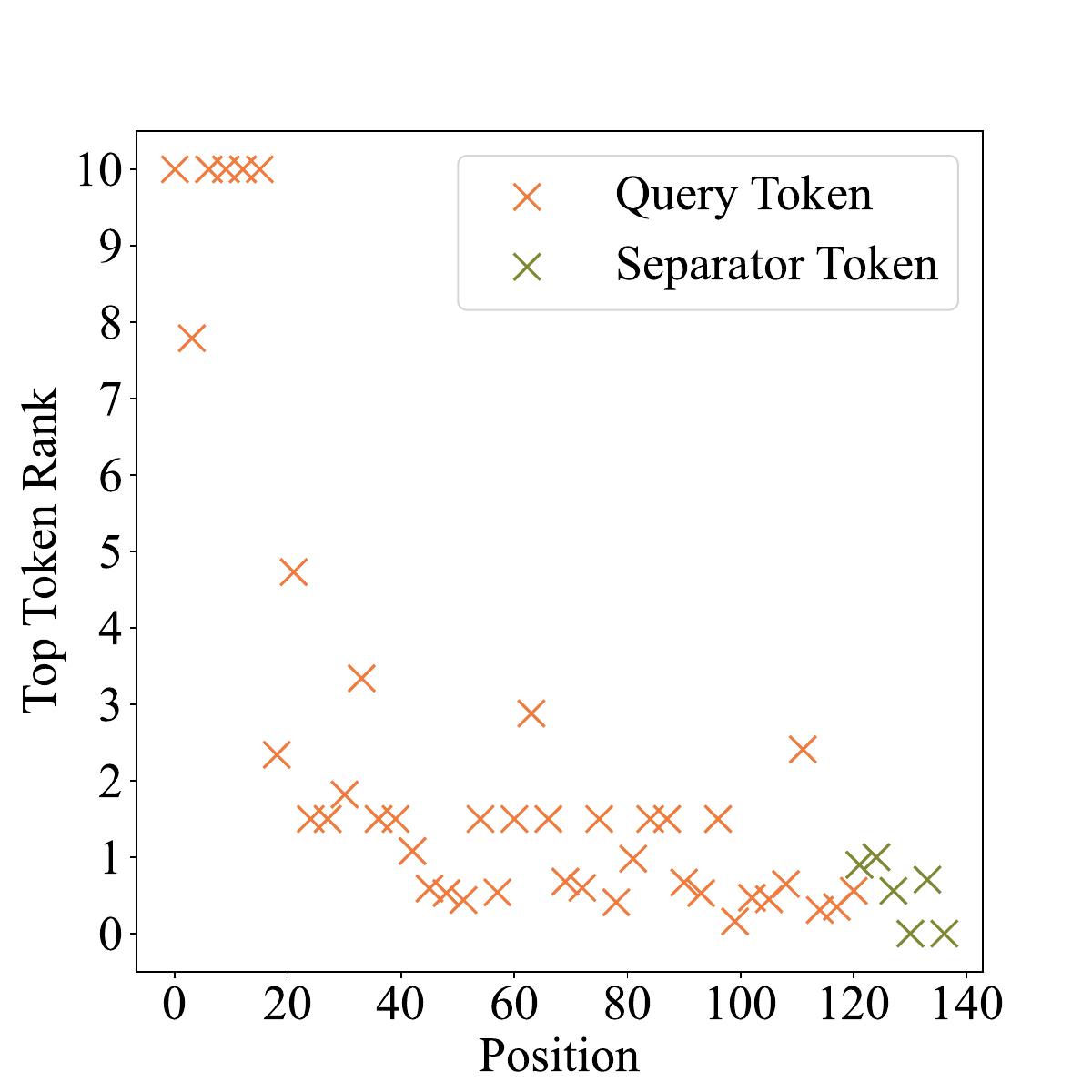}
    \caption{Input Experimental Group}
  \end{subfigure}
  \begin{subfigure}[b]{0.24\linewidth}
    \includegraphics[width=\linewidth]{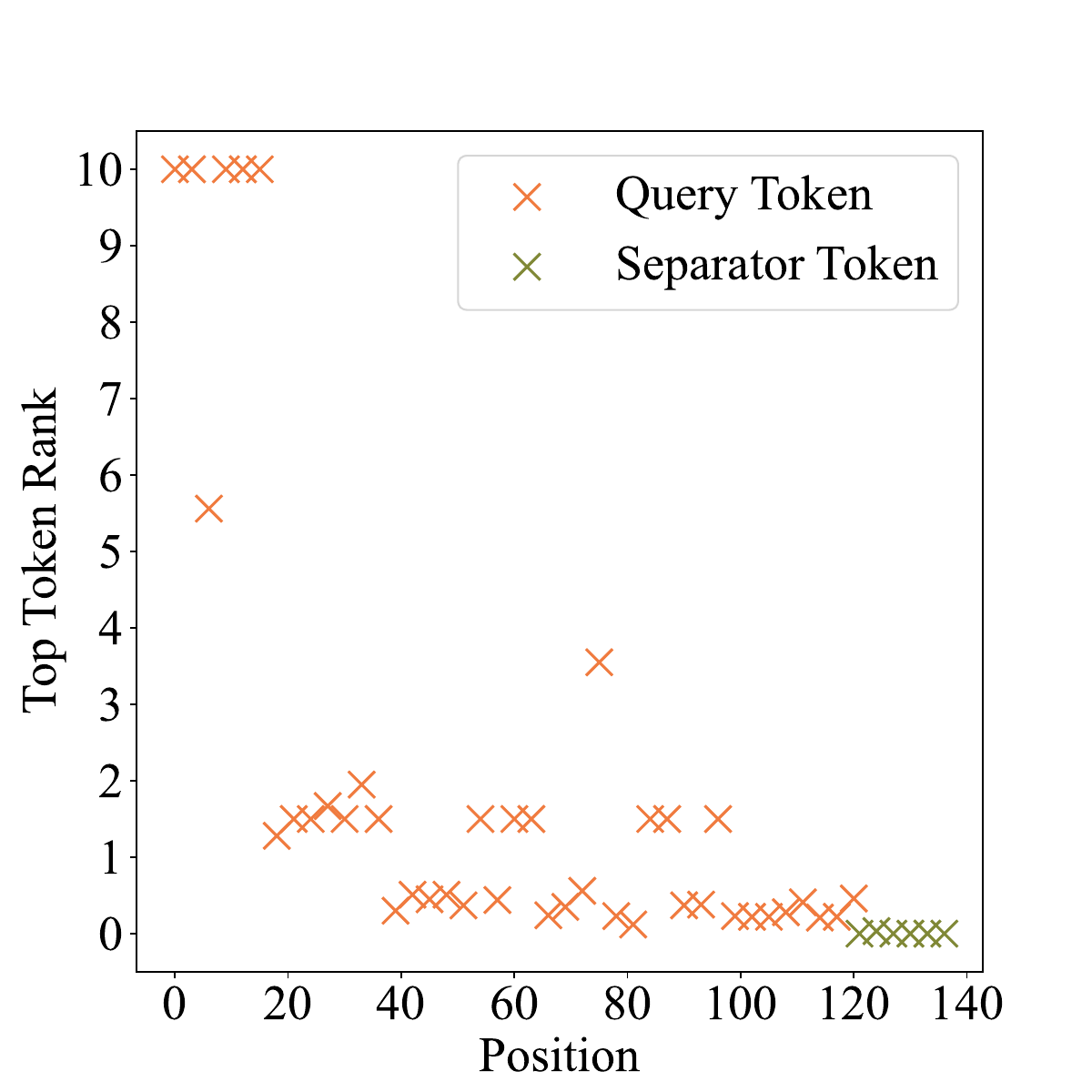}
    \caption{Input Control Group}
  \end{subfigure}
  \begin{subfigure}[b]{0.24\linewidth}
    \includegraphics[width=\linewidth]{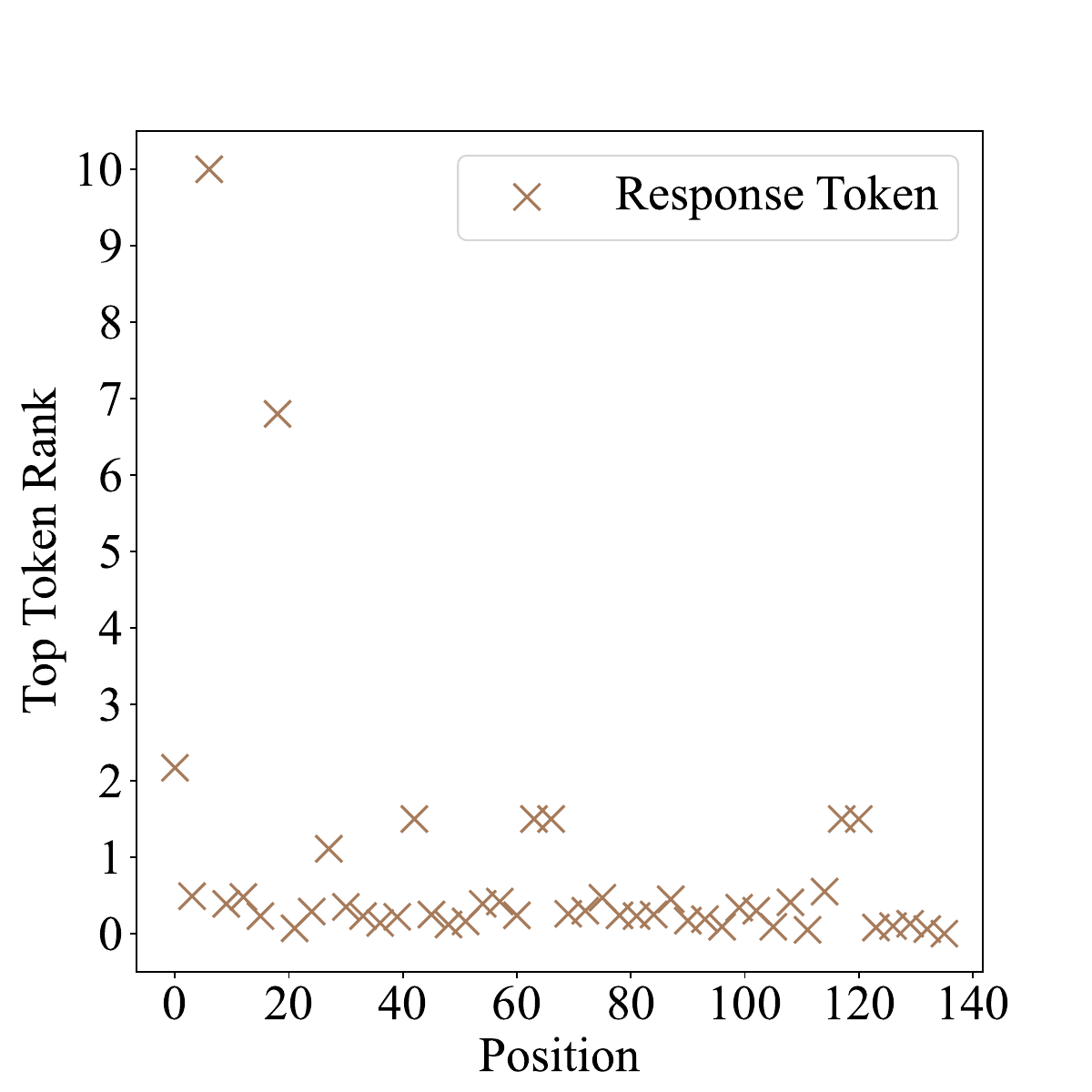}
    \caption{Output Experimental Group}
  \end{subfigure}
  \begin{subfigure}[b]{0.24\linewidth}
    \includegraphics[width=\linewidth]{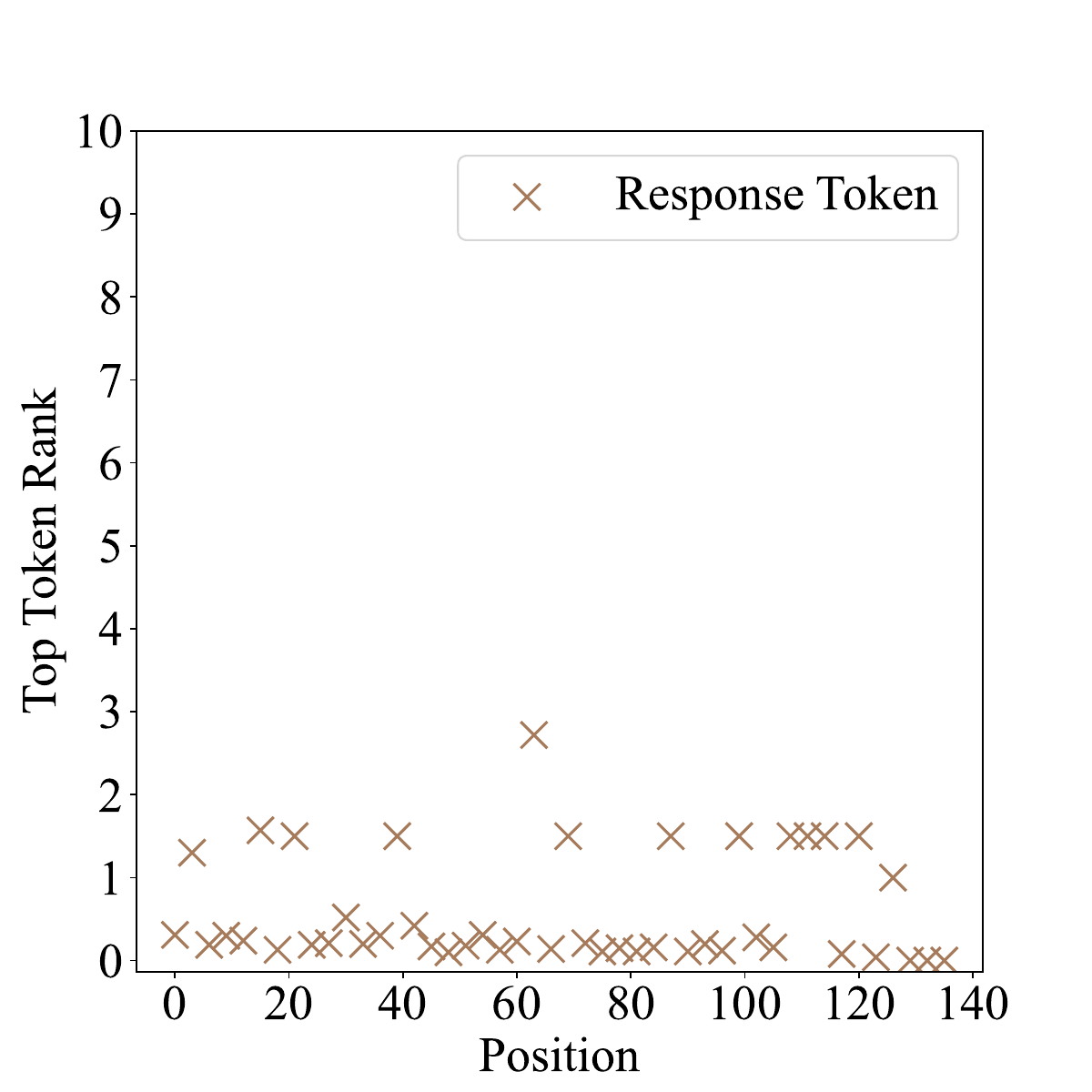}    
    \caption{Output Control Group}
  \end{subfigure} 

  \caption{Average Top Token Rank on Llama2-7b. \textit{Experimental Group} compares zero-shot and few-shot settings, while \textit{Control Group} compares two few-shot settings with different demonstrations. We visualize the input and output separately}
    \vspace{-0.4cm}
  \label{fig:ttr_llama}
\end{figure*}

\begin{figure*}[htbp!]
\centering
\begin{subfigure}[b]{0.24\linewidth}
    \includegraphics[width=\linewidth]{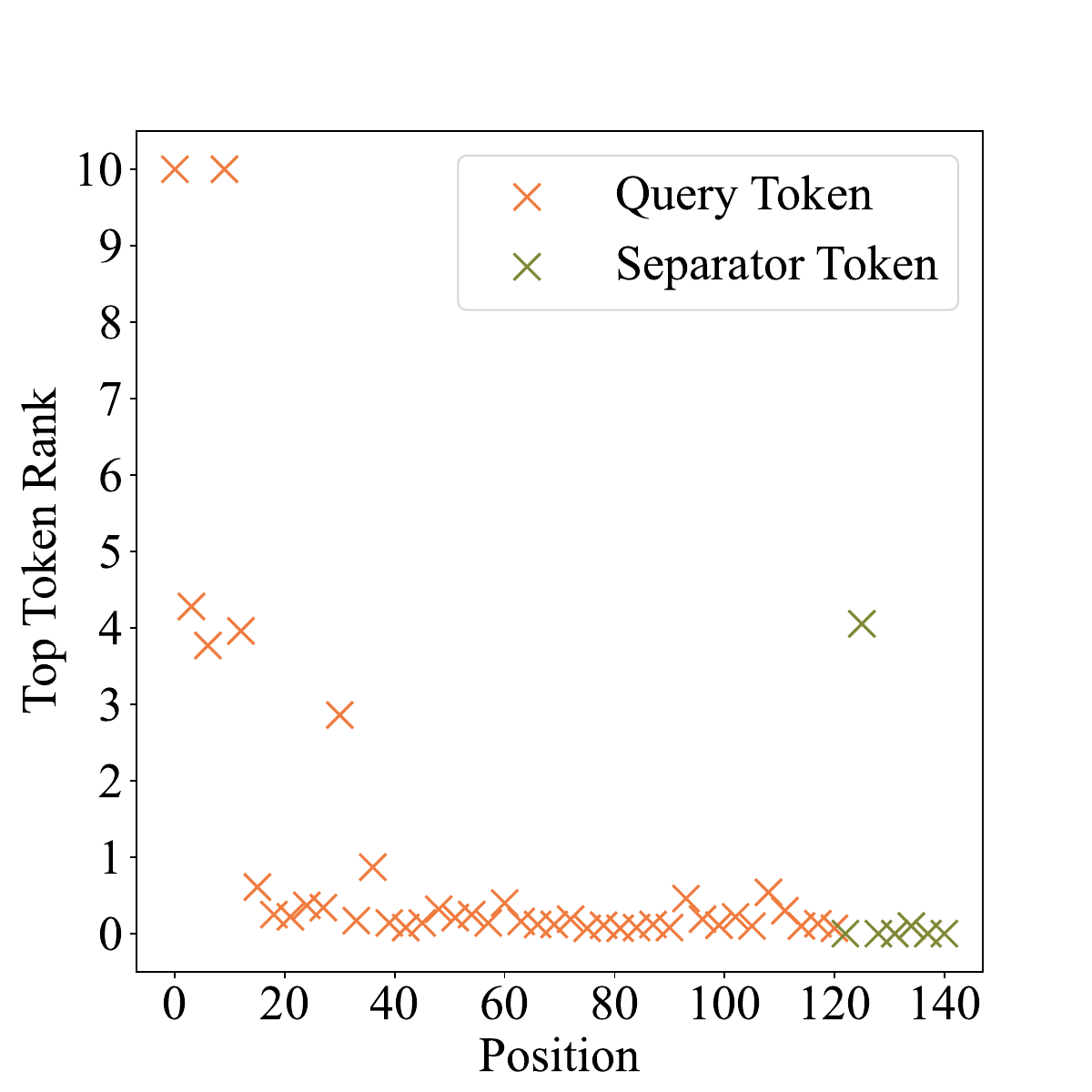}
    \caption{Input Experimental Group}
  \end{subfigure}
  \begin{subfigure}[b]{0.24\linewidth}
    \includegraphics[width=\linewidth]{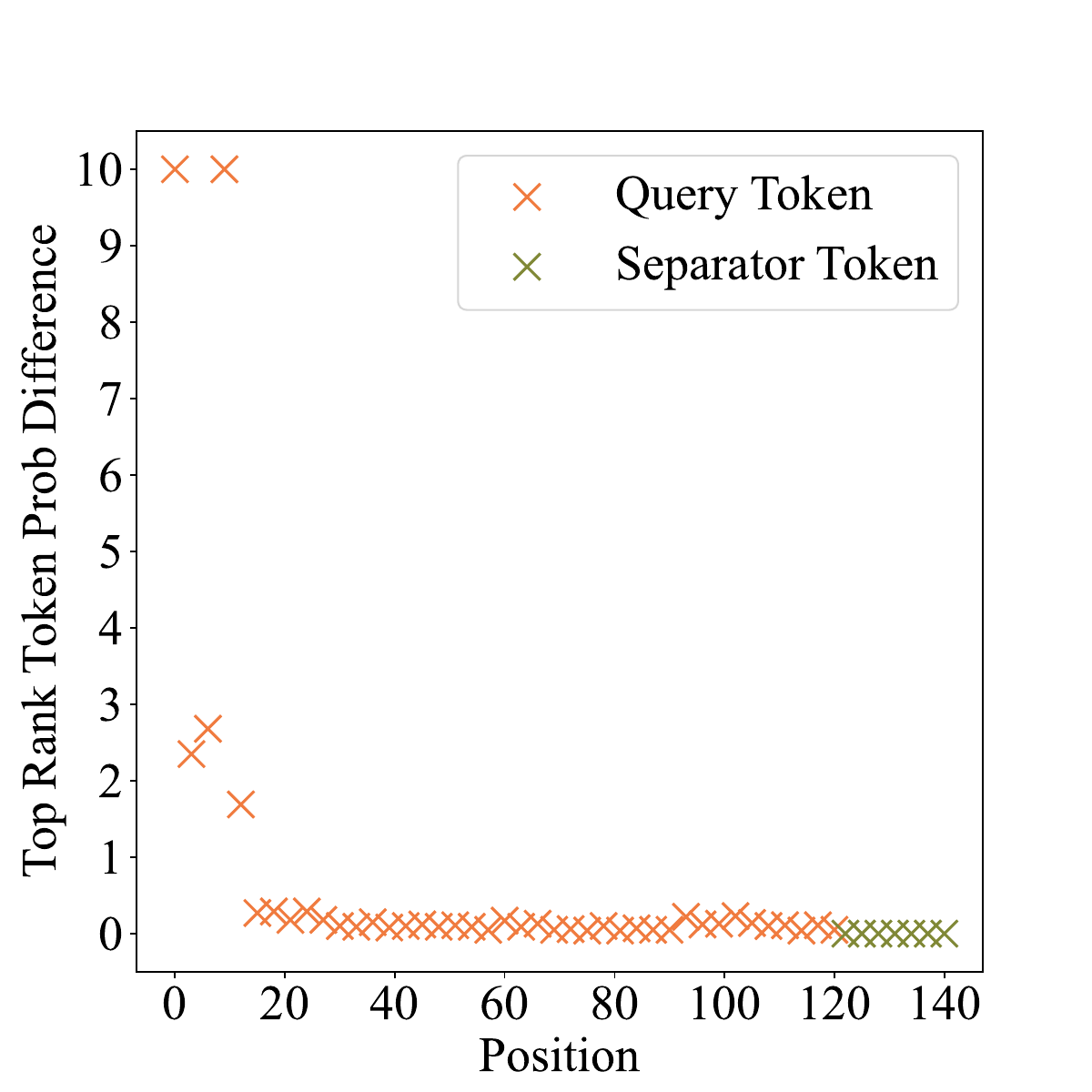}
    \caption{Input Control Group}
  \end{subfigure}
  \begin{subfigure}[b]{0.24\linewidth}
    \includegraphics[width=\linewidth]{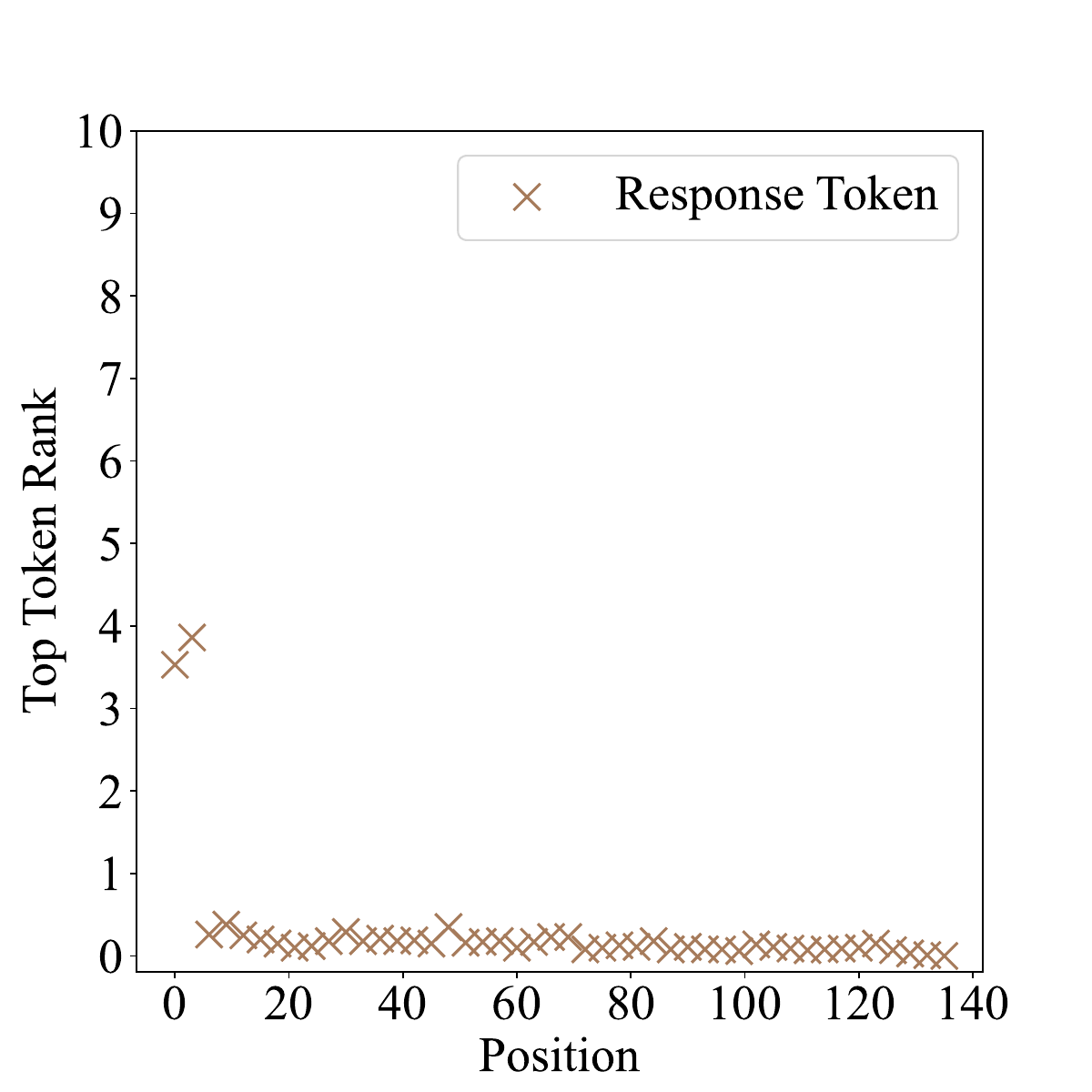}
    \caption{Output Experimental Group}
  \end{subfigure}
  \begin{subfigure}[b]{0.24\linewidth}
    \includegraphics[width=\linewidth]{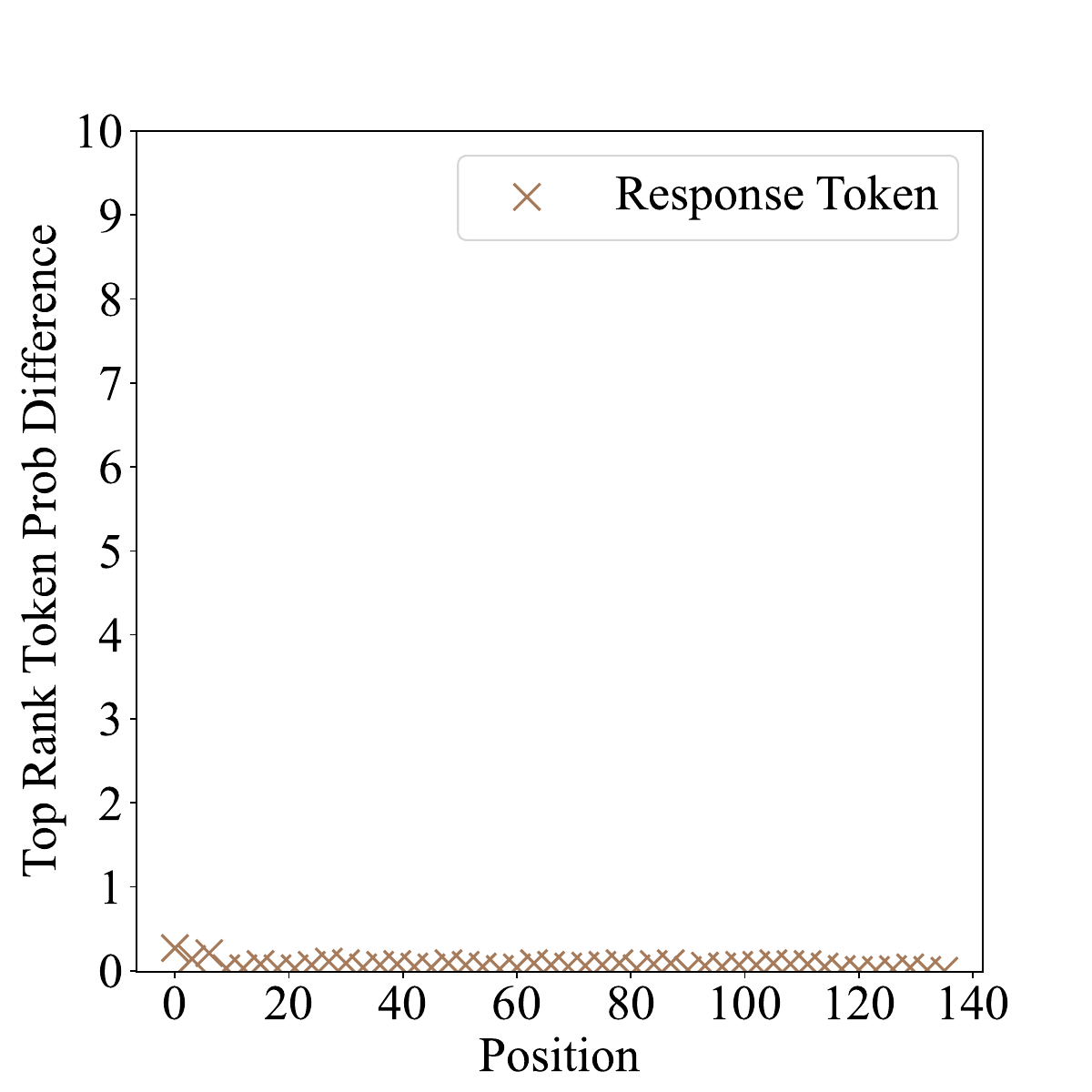}    
    \caption{Output Control Group}
  \end{subfigure} 

  \caption{Average Top Token Rank on Mistral-7b. \textit{Experimental Group} compares zero-shot and few-shot settings, while \textit{Control Group} compares two few-shot settings with different demonstrations. We visualize the input and output separately}
    \vspace{-0.4cm}
  \label{fig:ttr_mistral}
\end{figure*}

\begin{figure*}[htbp!]
\centering
\begin{subfigure}[b]{0.24\linewidth}
    \includegraphics[width=\linewidth]{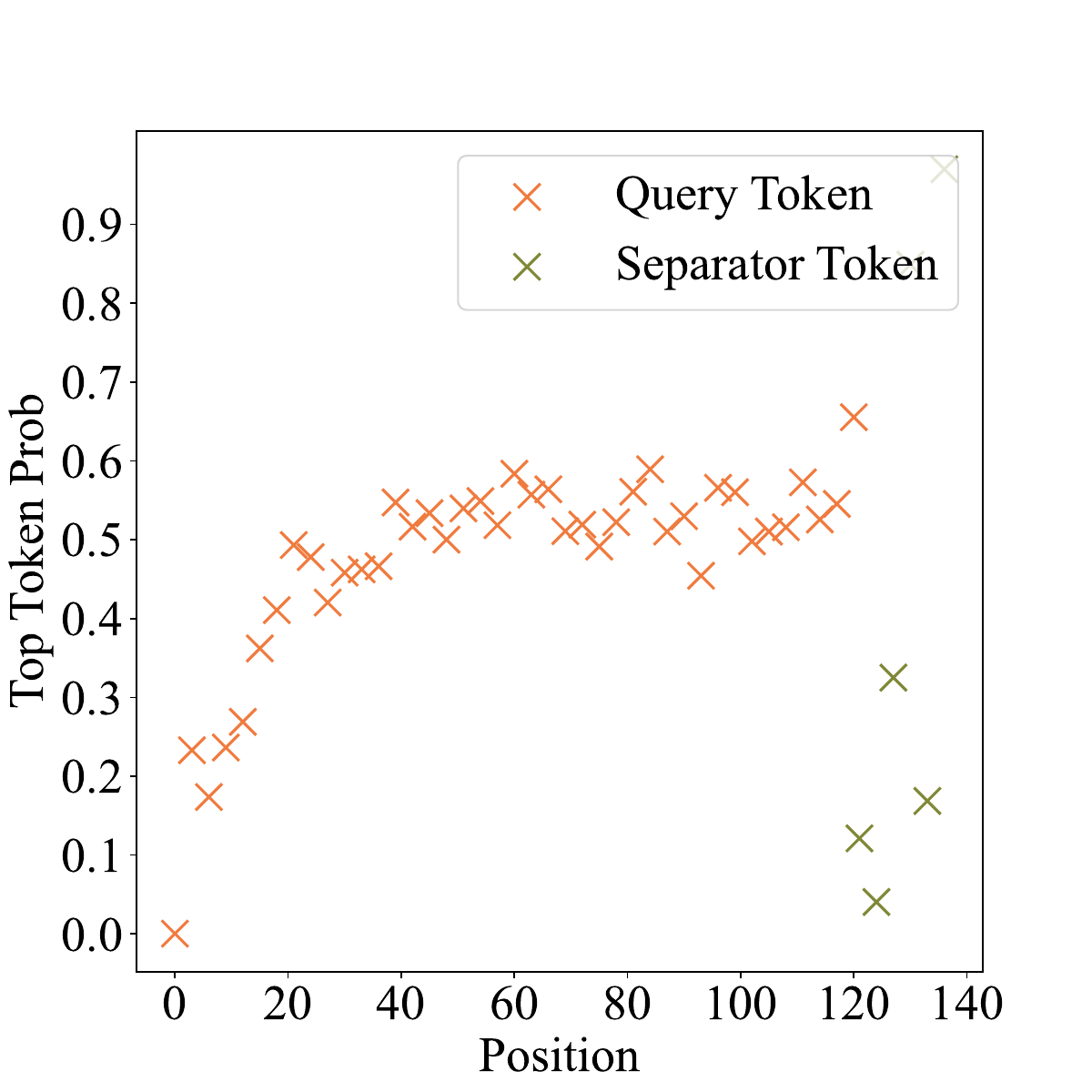}
    \caption{Input Experimental Group}
  \end{subfigure}
  \begin{subfigure}[b]{0.24\linewidth}
    \includegraphics[width=\linewidth]{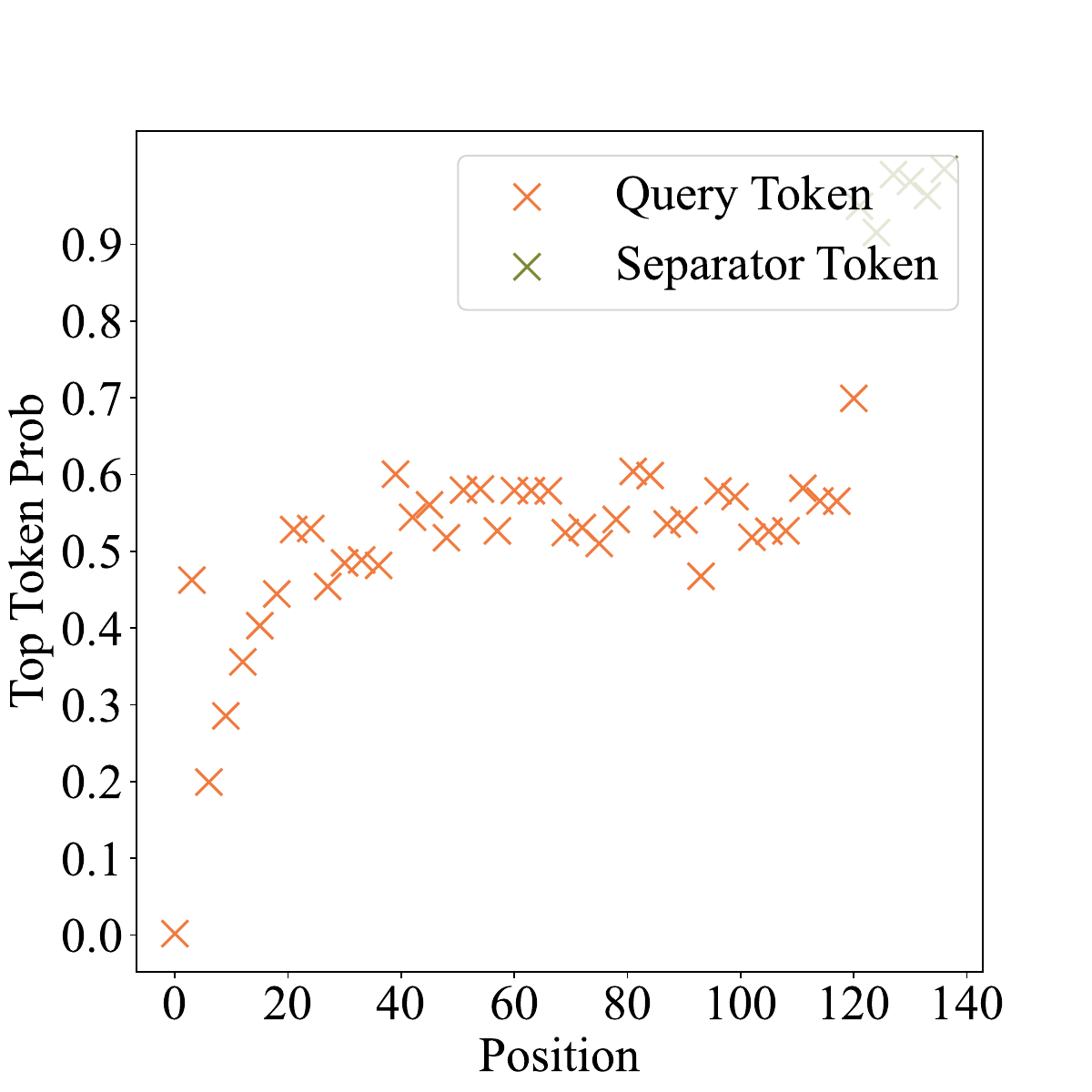}
    \caption{Input Control Group}
  \end{subfigure}
  \begin{subfigure}[b]{0.24\linewidth}
    \includegraphics[width=\linewidth]{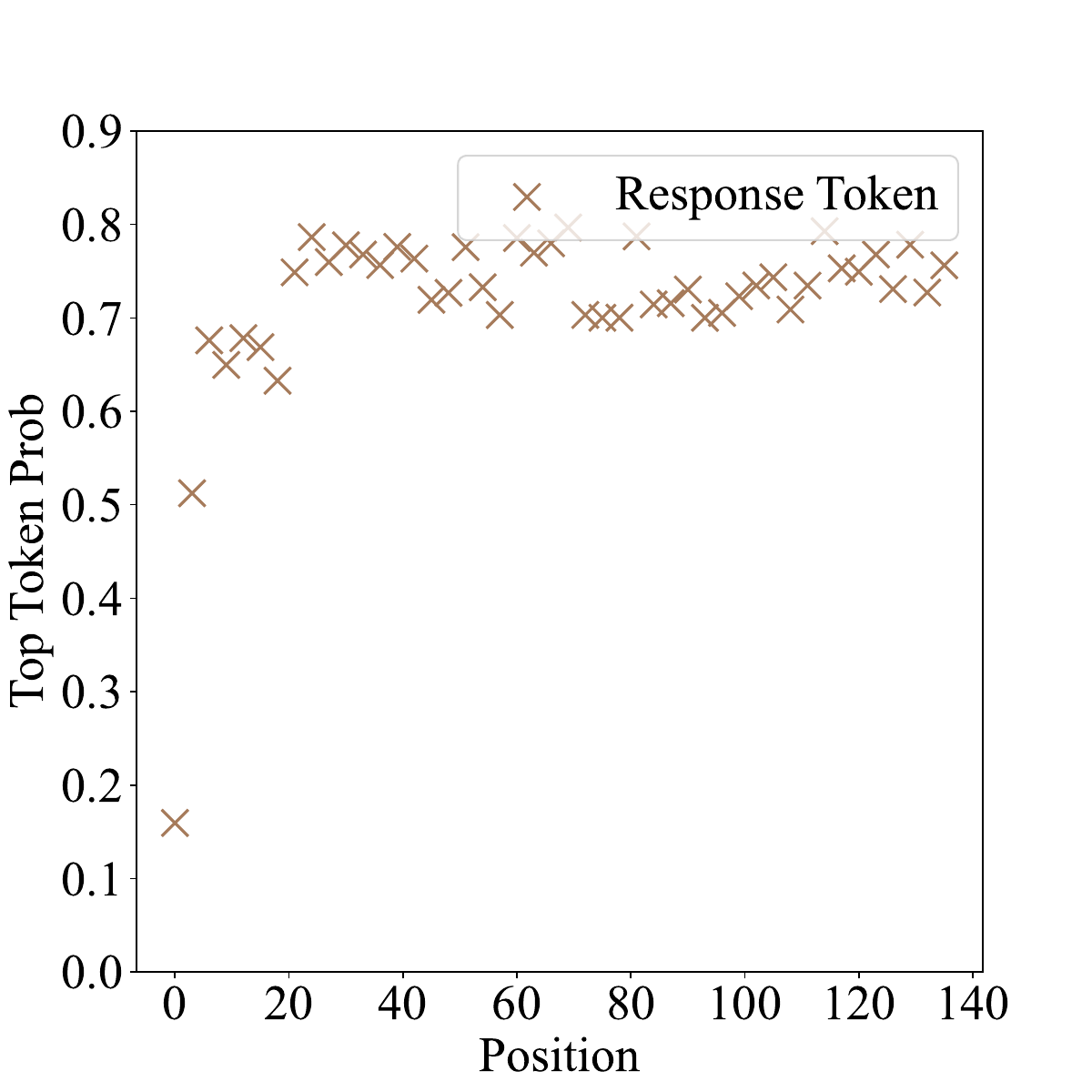}
    \caption{Output Experimental Group}
  \end{subfigure}
  \begin{subfigure}[b]{0.24\linewidth}
    \includegraphics[width=\linewidth]{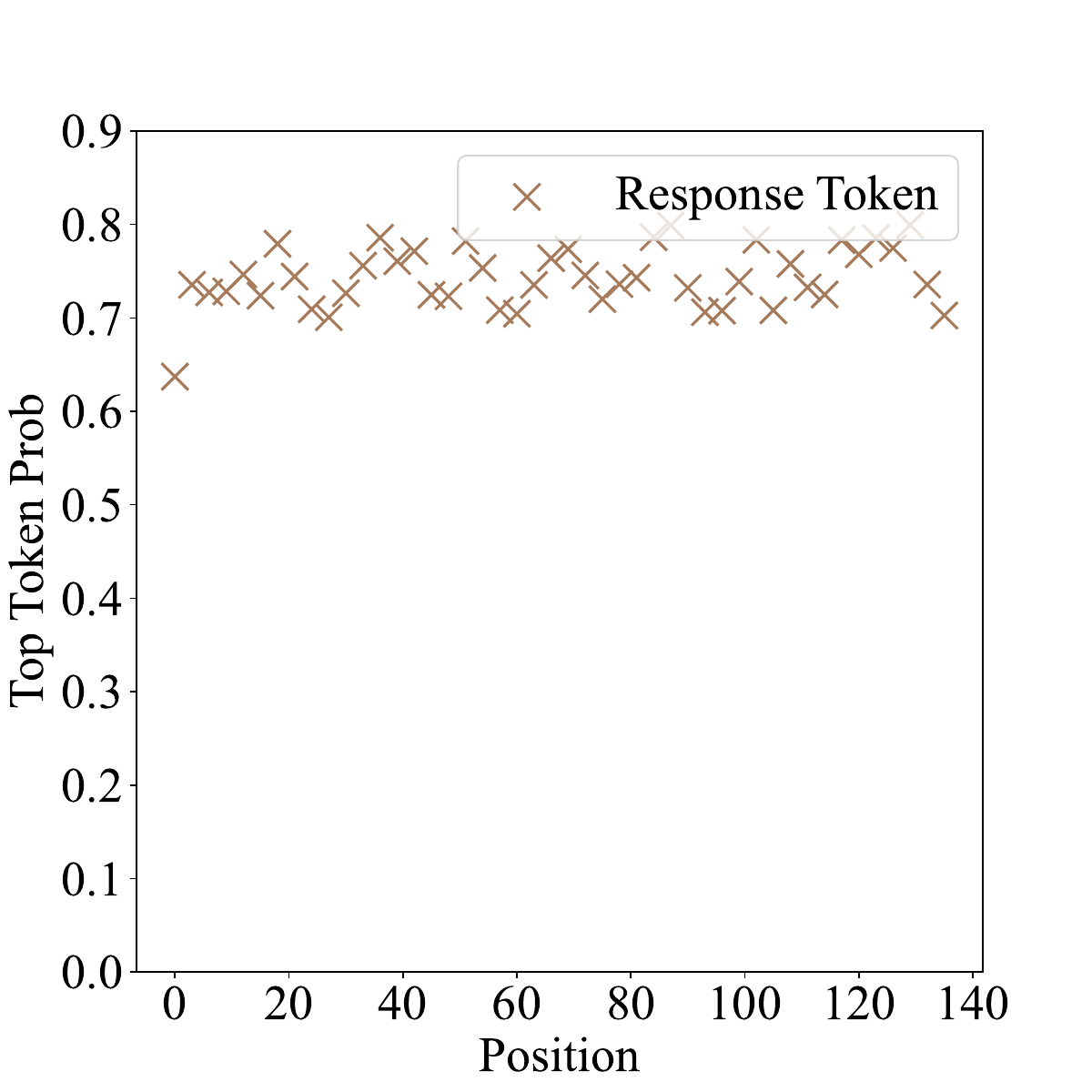}    
    \caption{Output Control Group}
  \end{subfigure} 

  \caption{Average  Top Token Prob on Llama2-7b. \textit{Experimental Group} compares zero-shot and few-shot settings, while \textit{Control Group} compares two few-shot settings with different demonstrations. We visualize the input and output separately}
    \vspace{-0.4cm}
  \label{fig:ttp_llama}
\end{figure*}

\begin{figure*}[htbp!]
\centering
\begin{subfigure}[b]{0.24\linewidth}
    \includegraphics[width=\linewidth]{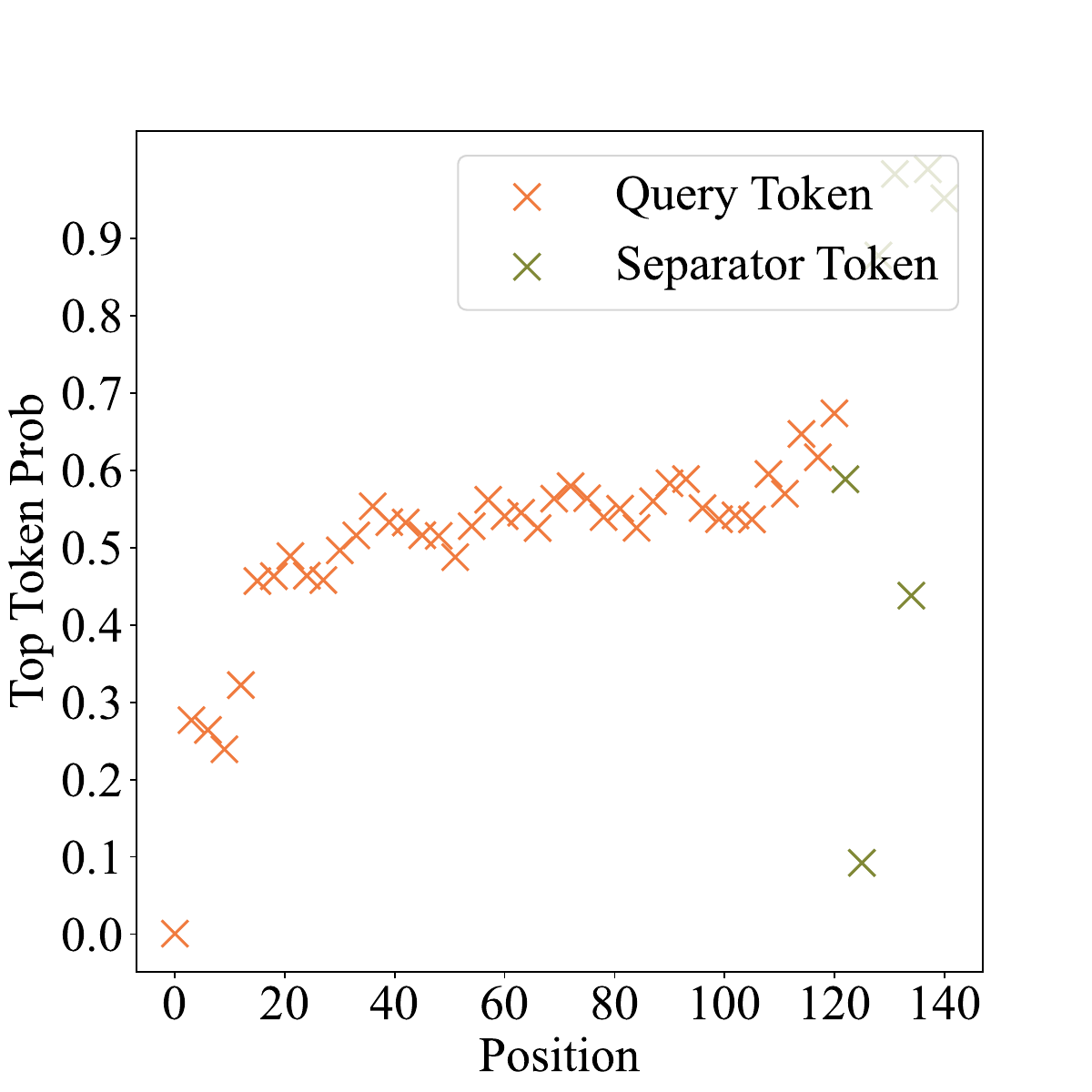}
    \caption{Input Experimental Group}
  \end{subfigure}
  \begin{subfigure}[b]{0.24\linewidth}
    \includegraphics[width=\linewidth]{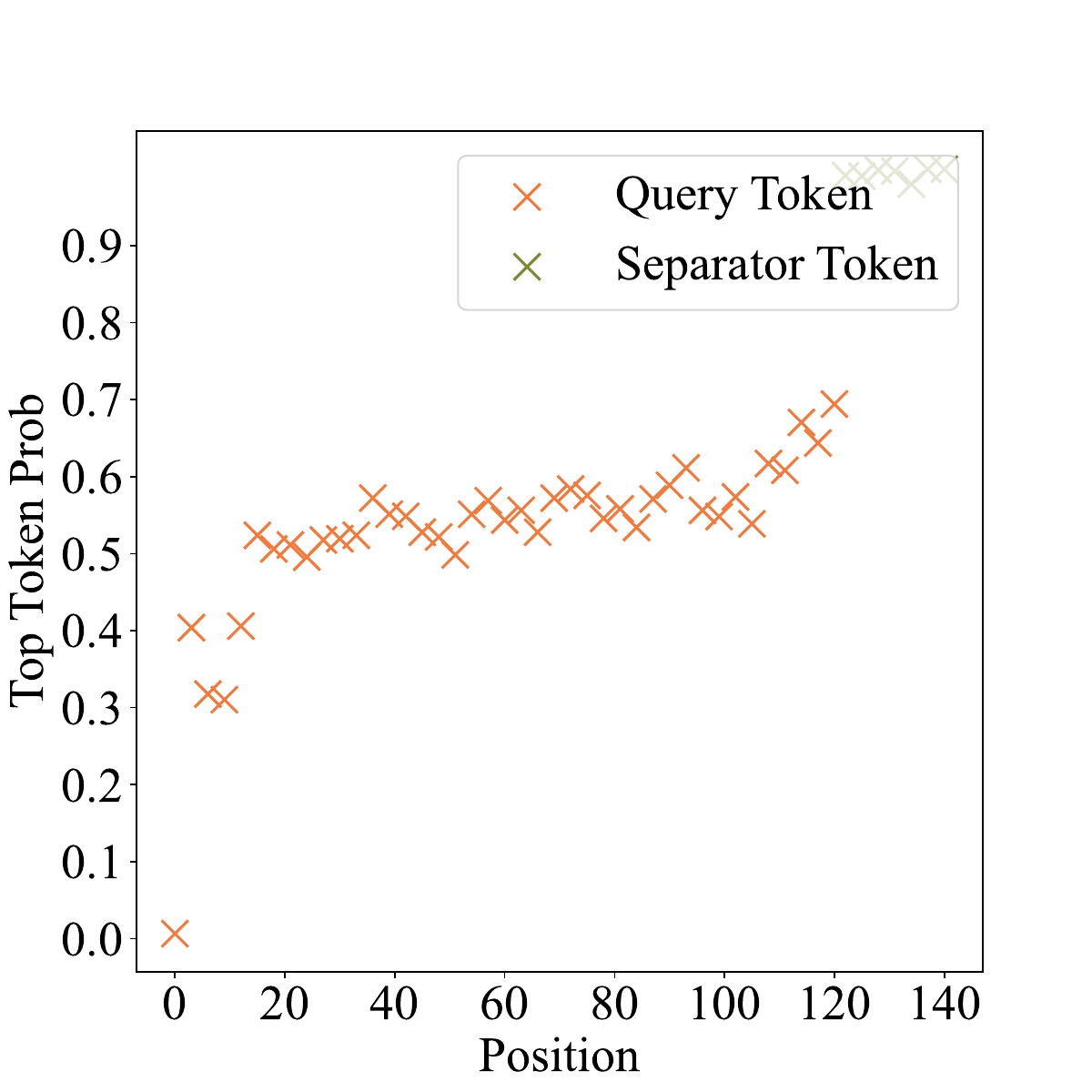}
    \caption{Input Control Group}
  \end{subfigure}
  \begin{subfigure}[b]{0.24\linewidth}
    \includegraphics[width=\linewidth]{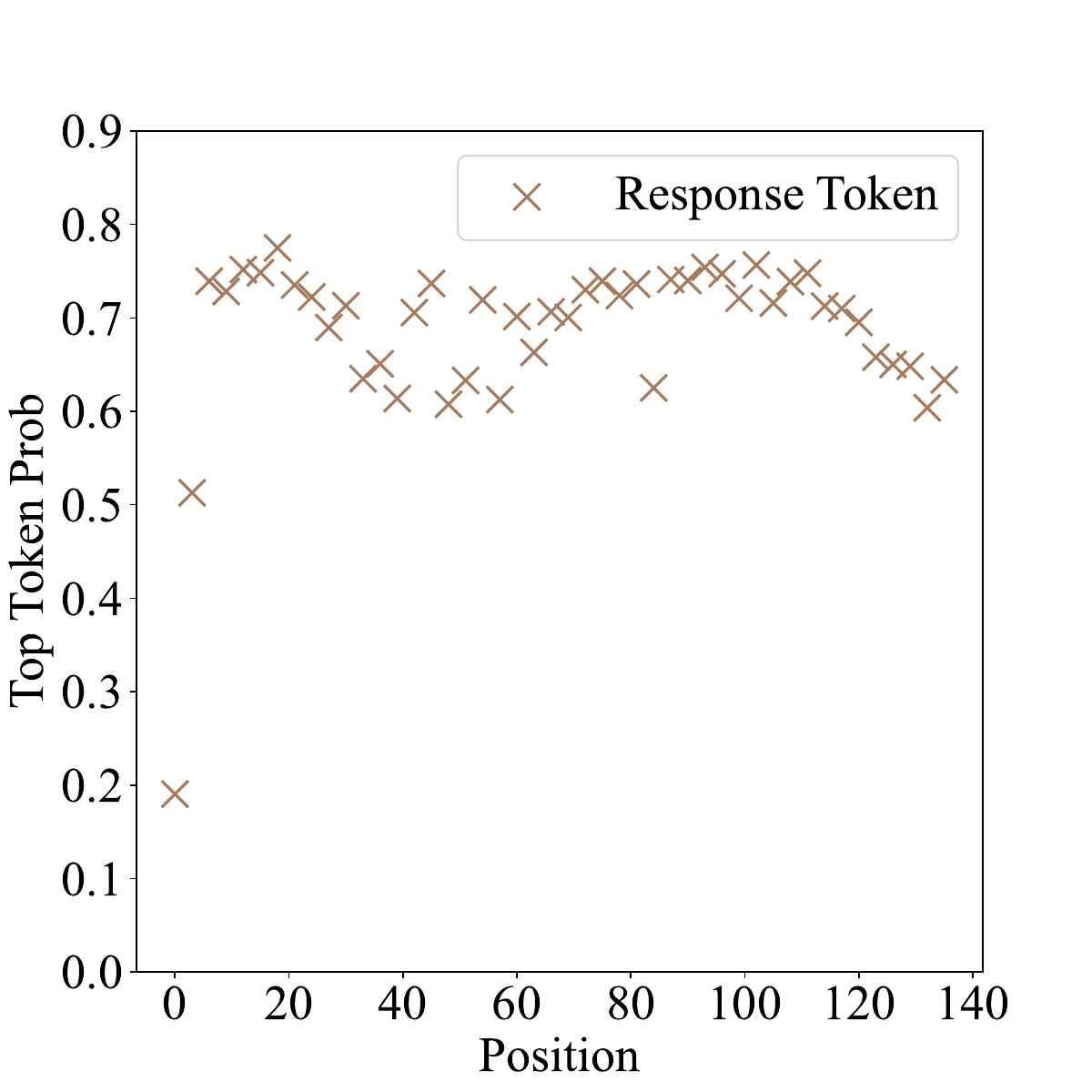}
    \caption{Output Experimental Group}
  \end{subfigure}
  \begin{subfigure}[b]{0.24\linewidth}
    \includegraphics[width=\linewidth]{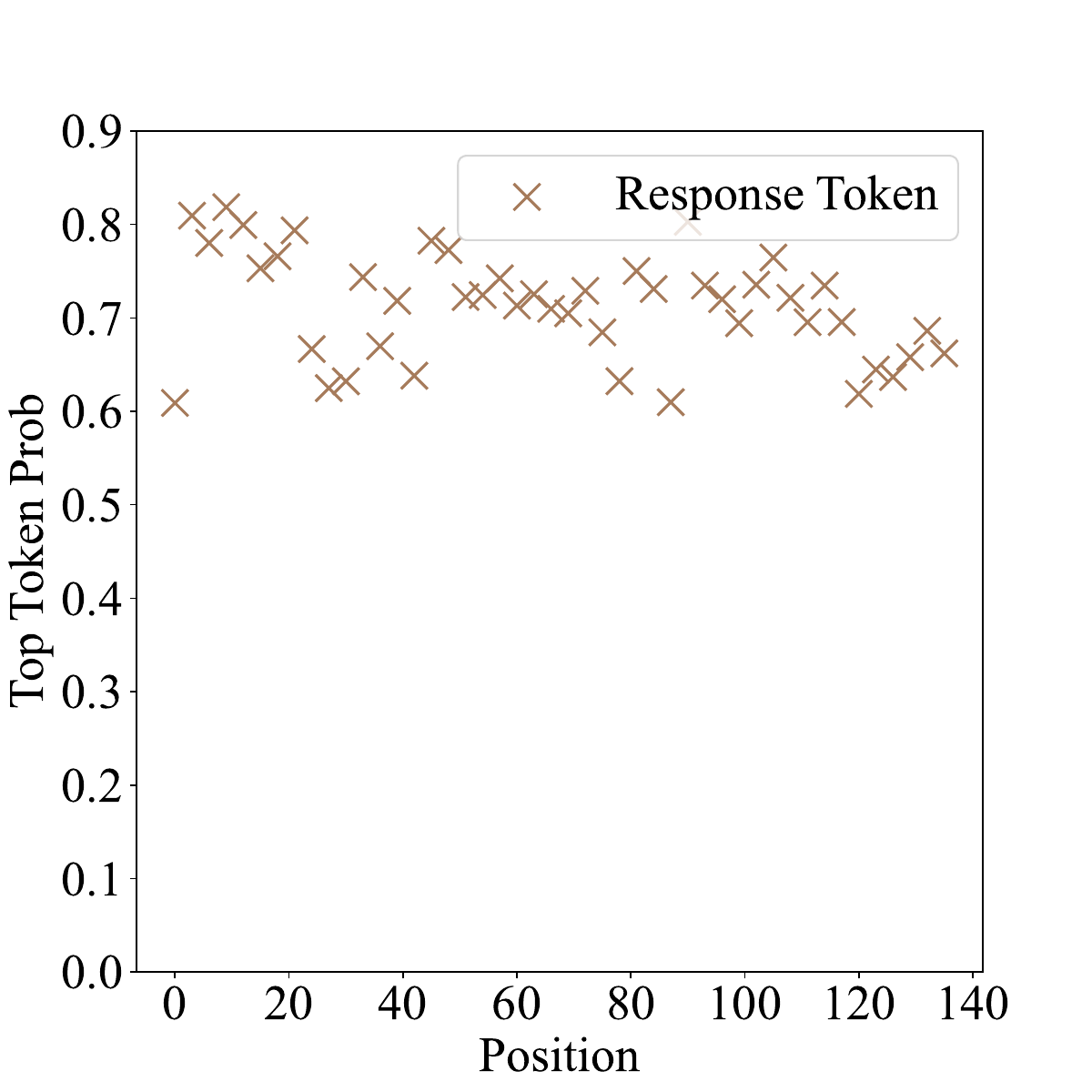}    
    \caption{Output Control Group}
  \end{subfigure} 

  \caption{Average Top Token Prob on Mistral-7b. \textit{Experimental Group} compares zero-shot and few-shot settings, while \textit{Control Group} compares two few-shot settings with different demonstrations. We visualize the input and output separately}
    \vspace{0.2cm}
  \label{fig:ttp_mistral}
\end{figure*}




%
We present additional comparative experiments to further delve into the impact of demonstrations on token representation. We conduct experiments on Llama2-7b and Mistral-7b models with the same data as in \S\ref{sec:motivation}. The experimental group includes both zero-shot and few-shot methods, while the control group includes two different demonstrations in few-shot settings. In addition to the KL-divergence of token distributions, we introduce two new metrics for measuring the difference between the two methods, i.e., Top Token Prob and Top Token Rank.

Top Token Rank refers to the ranking position of a token predicted by one method within the token distribution of another method. Specifically, given the context, we first obtain the next predicted token from one method and then determine its rank within the token distribution of the other method. A lower Top Token Rank manifests a greater overlap in the decision space under the greedy search setting.
Analogously, Top Token Prob indicates the probability of a token predicted by one method within the distribution of another method. In this case, we first obtain the next predicted token from one method and then report its probability in the token distribution of the other method. A higher Top Token Prob manifests a higher likelihood of obtaining the same result under the greedy sampling setting.
These metrics directly reflect the similarity between the generation results of the two methods. By evaluating both Top Token Rank and Top Token Prob, we can obtain a comprehensive understanding of how closely the methods align in terms of their token predictions.
The KL-divergence results on Mistral-7b are shown in \autoref{fig:kl_score_mixtral}, and we find a similar pattern to that in \autoref{fig:kl_score}.

The Top Token Rank results are shown in \autoref{fig:ttr_llama} and \autoref{fig:ttr_mistral}. For convenience, we set the rank of the token to 10 if it is greater than 10. We find that tokens with a large KL-divergence difference typically also have a higher Top Token Rank, indicating that our understanding of demonstrations applies to this observation as well. Notably, even though the separator token distribution differs significantly, the Top Token Rank remains low. This observation suggests that though demonstrations have a lot of influence on the separator token representation, the predicted next token rank still remains unchanged.

The Top Token Prob results are shown in \autoref{fig:ttp_llama} and \autoref{fig:ttp_mistral}, where we find that tokens with a large KL-divergence difference typically also have a low Top Token Prob. This further supports our understanding of the role that demonstrations play in the ICL. Similar to the result of Top Token Rank, the predicted separator token probability is high, indicating that demonstration will not change the selection of separator token. 

Overall, we observe similar patterns across KL-Divergence, Top Token Rank, and Top Token Prob metrics, despite minor differences. This demonstrates the generalizability and universality of our understanding of the impact of demonstrations.

\section{PICA Prompt}
\begin{table*}[h]
    \centering
    \renewcommand{\arraystretch}{1.1} 
    \scalebox{0.9}{
        \begin{tcolorbox}
                \textbf{The default version of PICA prompt with an example} 
                \tcblower
                \footnotesize
                \# Instruction \\
                \\
                Below is a list of conversations between a human and an AI assistant (you). \\
                As an AI assistant, you will engage in conversations with users, responding to their queries which are presented under the heading "\# Query:". \\
                Your responses should be entered under the heading "\# Answer:". \\
                You excel in a wide range of tasks including, but not limited to, providing general information, conducting reasoning, engaging in role-play, creative writing, planning, and solving mathematical and coding problems. \\
                Your responses should be well-structured, comprehensive, and aim to thoroughly address the user's query or problem at hand. \\
                When enumerating items in your responses, limit the examples to no more than ten, and avoid completely redundant content. \\
                Please ensure that your responses are encapsulated within triple backticks (``\textasciigrave \textasciigrave \textasciigrave'') at the start and end to maintain formatting consistency throughout the conversation. \\
                \\
                \# Query: \\
                \textasciigrave \textasciigrave \textasciigrave \\
                Find poems that mention the moon, including the poem titles and their poets. \\
                \textasciigrave \textasciigrave \textasciigrave \\
                \\
                \# Answer: \\
                \textasciigrave \textasciigrave \textasciigrave \\
                These are some examples of poems that mention the moon. \\
                1. "The Moon and the Yew Tree" by Sylvia Plath \\
                2. "The Moon" by Robert Louis Stevenson \\
                3. "Above the Dock" by T. E. Hulme (...)\\
                \textasciigrave \textasciigrave \textasciigrave 
        \end{tcolorbox}
    }
    \caption{The default version of PICA prompt with an example}
    \label{tab:full_prompt}
\end{table*}

We present the default version prompt with one example used in our experiment in the~\autoref{tab:full_prompt}. 

\section{Additional Experiment}

\begin{table*}[ht!]
\centering
\small
{
\setlength{\tabcolsep}{4pt}
	\begin{tabular}{l|cc|ccc ccc|c}
		\toprule
        \multirow{2}{*}[-2.5pt]{\textbf{Models + Alignment Methods}}       &\multicolumn{2}{c|}{\textbf{Alpaca-eval}}    &\multicolumn{6}{c|}{\textbf{Just-eval}} &\multirow{2}{*}[-2.5pt]{\textbf{Speedup}}  \\ 
        \cline{2-9}
		                          &\multicolumn{1}{c}{vs GPT-3} &\multicolumn{1}{c|}{vs GPT-4}   &\multicolumn{1}{c}{Helpful}    &\multicolumn{1}{c}{Clear}      &\multicolumn{1}{c}{Factual}    &\multicolumn{1}{c}{Deep}   &\multicolumn{1}{c}{Engaging}   &\multicolumn{1}{c|}{Safe}  \\
        \hline
        GPT-4-0613 	                     &72.51 	&53.52 	&4.86 	&4.99 	&4.90 	&4.49 	&4.61 	&4.97 	- \\
        \hline\hline                
        Llama2-70b-chat (RLHF)           &69.58     &47.19 	&4.90 	&4.96 	&4.88 	&4.72 	&4.80   &5.00 	&6.92  \\
        Llama2-70b (Zero-shot)           &24.65     &11.74  &2.78   &3.01   &3.11   &2.27   &2.29   &1.05   &5.81  \\
        Llama2-70b (Vanilla ICL)         &66.03 	&44.18 	&4.83 	&4.89 	&4.78 	&4.52 	&4.56 	&4.71 	&1.00   \\
        \rowcolor{lightgray} 
        Llama2-70b (Vec.)                &58.63 	&35.47 	&4.72 	&4.79 	&4.71 	&4.15 	&4.25 	&3.71 	&6.81   \\
        \rowcolor{lightgray} 
        Llama2-70b (Prog.)               &63.13 	&42.61 	&4.79 	&4.85 	&4.76 	&4.19 	&4.31 	&4.68 	&6.70   \\
        \rowcolor{lightgray} 
        Llama2-70b (PICA)                &68.66 	&45.31 	&4.85 	&4.85 	&4.82 	&4.21 	&4.58 	&4.70 	&6.73   \\
        \hline\hline
	\end{tabular}
 }
    \caption{Comparison of alignment performance and efficiency on Llama2-70b. Alpaca-eval presents the win rate against competitor models, while Just-eval presents the scores across six aspects (scores are on a scale of 1-5). Results highlighted in gray represent our methods: \textit{Vec.} denotes the ICL vector guidance and  \textit{Prog.} denotes progressive generation ablation variants. Speedup indicates the efficiency improvement compared to vanilla ICL.}
    \label{tab: additional alignment result}
    \vspace{-0.4cm}
\end{table*}

In this section, we present the performance of PICA on the larger Llama2-70B model. As shown in the table~\ref{tab: additional alignment result}, the PICA consistently achieves comparable results relative to vanilla ICL and RLHF methods. The observed performance improvements can be attributed to the proposed progressive generation strategy and ICL vector guidance. Notably, when implemented on the Llama2-70B model, PICA achieves 95\% of GPT-4's performance. These results demonstrate that the PICA remains effective even in larger model configurations.

\end{document}